\documentclass{article}

\usepackage{url}
\usepackage{graphicx}
\graphicspath{{figures/}}
\DeclareGraphicsExtensions{.eps}
\usepackage[caption=false,font=footnotesize]{subfig}
\usepackage{booktabs} 
\usepackage{amsmath, amssymb, amsthm}
\usepackage[noend]{algorithmic}
\usepackage[ruled, vlined, linesnumbered,algo2e]{algorithm2e}
\usepackage{color}

\newtheorem{theorem}{Theorem}
\newtheorem{definition}{Definition}

\newtheorem{lemma}{Lemma}

\usepackage[accepted]{icml2021}

\icmltitlerunning{Active Deep Learning on Entity Resolution by Risk Sampling}

\begin{document}

\twocolumn[
\icmltitle{Active Deep Learning on Entity Resolution by Risk Sampling}



\icmlsetsymbol{equal}{*}

\begin{icmlauthorlist}
\icmlauthor{Youcef Nafa}{to,goo}
\icmlauthor{Qun Chen}{to,goo}
\icmlauthor{Zhaoqiang Chen}{to,goo}
\icmlauthor{Xingyu Lu}{to,goo}
\icmlauthor{Haiyang He}{to,goo}
\icmlauthor{Tianyi Duan}{to,goo}
\icmlauthor{Zhanhuai Li}{to,goo}
\end{icmlauthorlist}

\icmlaffiliation{to}{School of Computer Science, Northwestern Polytechnical University, Xi’an, Shaanxi, China}
\icmlaffiliation{goo}{Key Laboratory of Big Data Storage and Management, Northwestern Polytechnical University, Xi’an, Shaanxi, China}

\icmlcorrespondingauthor{Youcef Nafa}{youcef.nafa@mail.nwpu.edu.cn}
\icmlcorrespondingauthor{Qun Chen}{chenbenben@nwpu.edu.cn}

\icmlkeywords{Active Learning, Deep Learning, Risk Analysis, Entity Resolution}

\vskip 0.3in
]



\printAffiliationsAndNotice{}  

\begin{abstract}
    While the state-of-the-art performance on entity resolution (ER) has been achieved by deep learning, its effectiveness depends on large quantities of accurately labeled training data. To alleviate the data labeling burden, Active Learning (AL) presents itself as a feasible solution that focuses on data deemed useful for model training.
    	
    Building upon the recent advances in risk analysis for ER, which can provide a more refined estimate on label misprediction risk than the simpler classifier outputs, we propose a novel AL approach of risk sampling for ER. Risk sampling leverages misprediction risk estimation for active instance selection. Based on the core-set characterization for AL, we theoretically derive an optimization model which aims to minimize core-set loss with non-uniform Lipschitz continuity. Since the defined weighted K-medoids problem is NP-hard, we then present an efficient heuristic algorithm. Finally, we empirically verify the efficacy of the proposed approach on real data by a comparative study. Our extensive experiments have shown that it outperforms the existing alternatives by considerable margins. Using ER as a test case, we demonstrate that risk sampling is a promising approach potentially applicable to other challenging classification tasks.
\end{abstract}

\section{Introduction}

\begin{figure}[!t]
    \centering
    \subfloat{\includegraphics[width=0.9\linewidth]{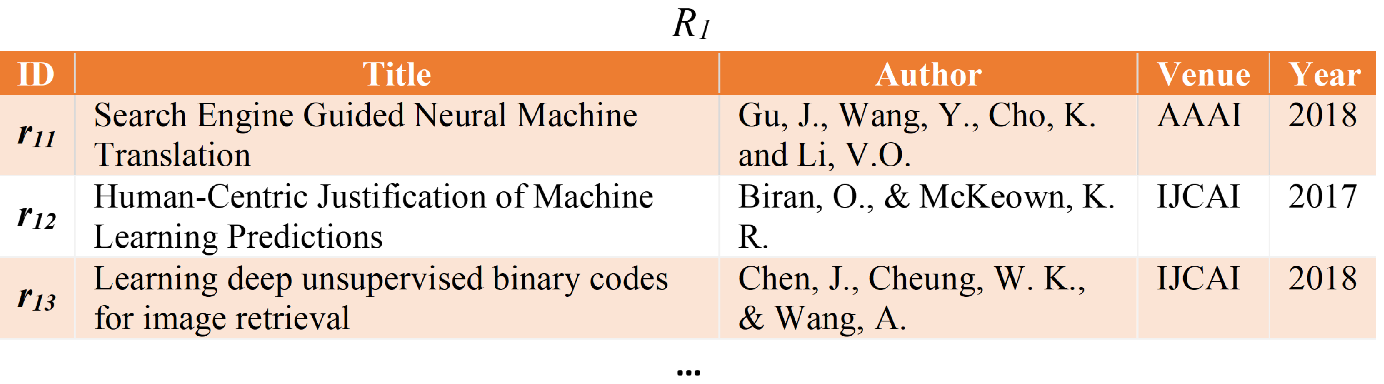}} \\
    \subfloat{\includegraphics[width=0.9\linewidth]{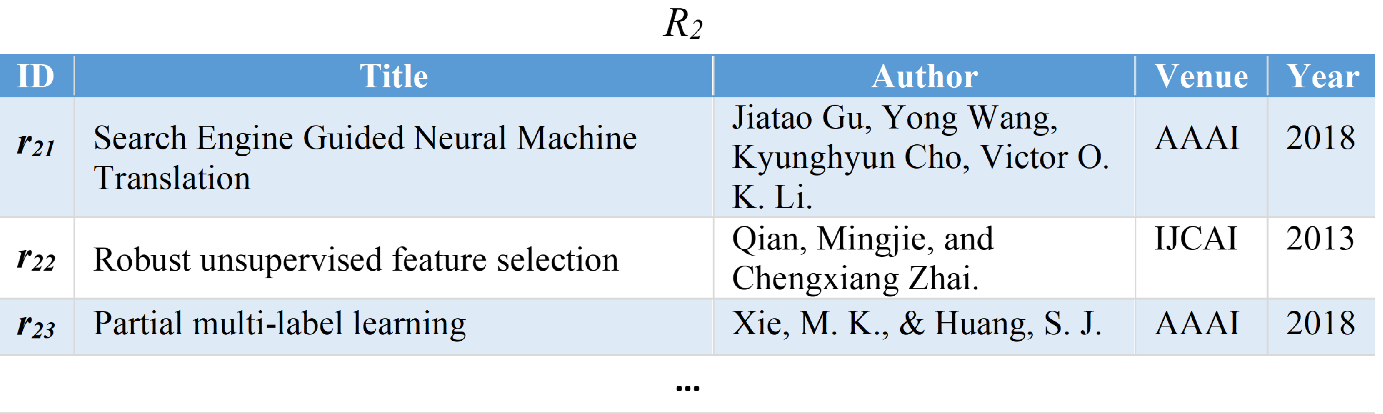}}
    \caption{An Entity Resolution running example. $R_1$ and $R_2$ are tables of paper records with the equivalent pair $\langle r_{11}, r_{21} \rangle$.}
    \label{fig:runningexample}
\end{figure}

The purpose of entity resolution (ER) is to identify the equivalent records that refer to the same real-world entity. Considering the running example shown in Fig.~\ref{fig:runningexample}, ER needs to match the paper records between two tables, $R_1$ and $R_2$. A pair $\langle r_{1i},r_{2j} \rangle$, in which $r_{1i}$ and $r_{2j}$ denote a record in $R_1$ and $R_2$ respectively, is called an \emph{equivalent} pair if and only if $r_{1i}$ and $r_{2j}$ refer to the same paper; otherwise, it is called an \emph{inequivalent} pair. In this example, $r_{11}$ and $r_{21}$ are \emph{equivalent} while $r_{11}$ and $r_{22}$ are \emph{inequivalent}. ER can be treated as a binary classification problem tasked with labeling record pairs as \emph{equivalent} or \emph{inequivalent}. Therefore, various learning models have been proposed for ER~\cite{svm08}. As many other classification tasks (e.g. image and speech recognition), the state-of-the-art performance on ER has been achieved by deep learning~\cite{deepmatcher,deepER,seq2seq,end2end,autoEM,bertEM}.
		
Unfortunately, the efficacy of Deep Neural Network (DNN) models depends on large quantities of accurately labeled training data, which may not be readily available in practical scenarios. One possible way to overcome this issue is by active learning, in which data are actively sampled to be labeled by human oracles with the goal of maximizing model performance while minimizing labeling costs. Various sampling strategies have been proposed for active learning over the years coming from different perspectives, e.g. uncertainty~\cite{uncertainty}, representativeness~\cite{coreset} and expected model change~\cite{EGLvoice}. There have also been different combinations of Uncertainty with Representativeness~\cite{kcover,elhamifar} or with Expected Model Change~\cite{EGLtext} in an attempt to get the best of both worlds. In the classical setting, active learning algorithms typically choose a single point at each iteration; however, this is not feasible for DNN models since i) a single point is likely to have no statistically significant impact on the
accuracy due to the locality of optimization methods, and ii) each iteration requires a full training until convergence which makes it intractable to query labels one-by-one. Hence, most proposed AL algorithms for DNNs~\cite{kcover, coreset, dpp, generative,lowres, badge}, take the strategy of batch selection that queries labels for a large subset at each iteration.

Uncertainty, considered the cheapest to obtain, is the mostly used sampling strategy due to its robustness across architectures and domains~\cite{benchmark}. Empirical studies~\cite{gissin2019discriminative} have also revealed that it is usually highly competitive with the existing but more complicated alternatives. We note that risk analysis for ER has been recently studied~\cite{VLDBWorkshop,r-HUMO,RISK} with the latter representing the most recent interpretable and learnable solution, henceforth denoted LearnRisk. Risk analysis estimates the misprediction risk of a classifier when applied to a certain workload. It has been empirically shown~\cite{RISK} that LearnRisk can identify mislabeled instances with considerably higher accuracy than the existing uncertainty measures, which are directly estimated upon classifier outputs. Since advanced risk analysis can provide a more refined estimate on label status uncertainty for unseen data, it is naturally fit as an AL strategy.
 
\begin{figure}[!t]
	\centering
	\includegraphics[width=0.9\linewidth]{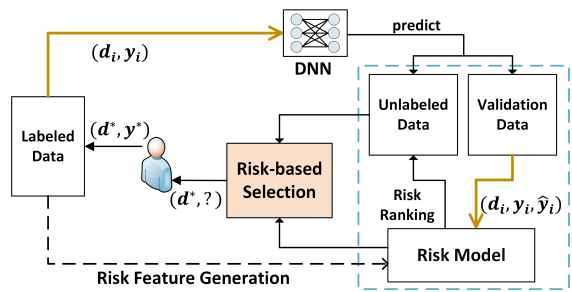}
	\caption{The Framework of Risk Sampling for Active Learning.}
	\label{fig:framework}
\end{figure}

Therefore, in this paper, we propose a novel AL approach of risk sampling for ER. Fig.~\ref{fig:framework} illustrates the risk sampling framework, which leverages the results of risk analysis in the sampling step of active learning. Based on the existing core-set characterization~\cite{coreset} for AL, we theoretically derive an optimization model which aims to minimize the core-set loss with non-uniform Lipschitz continuity. 
Since the defined weighted K-medoids problem is NP-hard, we then present an efficient heuristic algorithm. It is noteworthy that the strategy of risk sampling tends to sample challenging examples for the classifier, and spur it into learning better representations by looking farther than the low confidence regions in the current representation space. The main contributions of this paper can be summarized as follows:
\begin{itemize}
    \item We propose a novel approach of risk sampling for active learning. 
    \item We define a theoretically sound optimization model for risk sampling, and due to its NP-hardness, present an efficient weighted K-medoids algorithm.
    \item We present a technical AL solution for ER based on risk sampling.
    \item We verify the efficacy of the proposed solution through extensive experiments on real datasets. Our empirical study has shown that it can effectively accelerate classifier training compared with the existing alternatives. Furthermore, its performance is very robust w.r.t the size of validation data.
\end{itemize}

The rest of this paper is organized as follows: in Section~\ref{sec:related_works}, we discuss related work. We define the task and introduce the risk analysis technique in Section~\ref{sec:preliminaries}. In Section~\ref{sec:approach}, we propose the approach of risk sampling. Section~\ref{sec:experiments} describes the empirical evaluation results. Finally, Section~\ref{sec:conclusion} concludes this paper.

\section{Related Work}
\label{sec:related_works}

We review related work from the orthogonal perspectives of entity resolution and active learning.

\textbf{Entity Resolution.} 
Also known as Entity Matching or Record Linkage, ER plays a key role in data integration and has been extensively studied in the literature~\cite{christen2012data,web2015}. 
Other than the rule-based and probabilistic solutions~\cite{rules15,rules17,markov}, several machine learning models have been proposed, including Support Vector Machines~\cite{svm08}, end-to-end deep neural network architectures~\cite{deepmatcher,deepER,seq2seq,end2end}, and pretrained models~\cite{autoEM,bertEM}.

ER remains very challenging in real scenarios due to the prevalence of dirty data. Therefore, there is a need for \emph{risk analysis}, alternatively called trust scoring or confidence ranking in the literature. It encompasses a multitude of methods that all intend to detect situations where a deployed DNN model is prone to misprediction. The proposed solutions range from those simply based on the model's output probabilities to more sophisticated, interpretable, and learnable ones~\cite{baseline,trust,vision,RISK}. Among them, LearnRisk~\cite{RISK} is an interpretable and learnable framework for ER that is able to construct a dynamic risk model tuned towards a specific workload. It measures the risk using the VaR (Value-at-Risk)~\cite{tardivo2002value} metric from financial investment modeling. In this work we incorporate risk analysis into the standard pool-based active learning framework by treating misprediction risk as an informativeness measure.

\textbf{Active Learning.} Active learning has been extensively researched in the context of machine learning. The most prominent approaches that proved to perform well include {\it margin-based}, {\it maximum entropy}, {\it Query by committee} and {\it Expected variance reduction} to name a few~\cite{survey12} . However, many of the above methods pose challenges when applied to deep neural networks. The margin-based approaches are hindered by the fact that neural networks have an intractable decision boundary. Query by committee requires maintaining multiple classifiers and retraining them in each iteration which is not very practical. Similarly, the variance reduction methods require classifier retraining for each unlabeled point; this process is prohibitive even for shallow models. Active learning for ER has also received great attention~\cite{active02,active10,AL_ER_2020,largescale,vae_al_er}. In the low-resource setting, ER was also tackled using deep transfer active learning~\cite{lowres}.

More recently, active learning for DNNs has also been studied, mostly focusing on image classification. \cite{mc16,batchbald} showed that applying dropout at test time can approximate Bayesian inference enabling the application of Bayesian methods to deep learning. \cite{adversarial18} approximated the distance to decision boundary by the distance to the nearest adversarial example. \cite{EGLvoice} and~\cite{EGLtext} used an expected model change measure which chooses examples that maximize the impact on the learned model weights when labeled. Other recent works include generative data augmentation for AL~\cite{generative}, adversarial network-based discrimination of informative points~\cite{vae} and detrimental point processes-based batch selection~\cite{dpp} to name a few. There also exist proposals combining uncertainty with representativeness using data representation and entropy such as~\cite{kcover, elhamifar}, or relying on gradient-based representation and gradient amplitude as a proxy to uncertainty~\cite{badge}.
It is worthy to point out that the proposed approach of risk sampling can be easily generalized to image classification when provided with effective risk analysis techniques.

\section{Preliminaries} \label{sec:preliminaries}

In this section, we formally state the AL task, and then introduce the risk analysis technique for ER, LearnRisk.
	
\subsection{Task statement}
\label{sec:task}

Suppose that we have a set of record pairs $D=\{d_i, y_i\}$, where a pair $d_i$ can be labeled as \emph{equivalent} ($y_i = 1$) or \emph{inequivalent} ($y_i = 0$). We follow the standard pool-based setting in which the set of training data, $D$, is partitioned into a small initial labeled set $L=\{d_j,y_j\}$ and an unlabeled set $U$. We also assume the existence of two other sets: a validation set $V$ that is commonly used for hyperparameter tuning as well as early stopping for DNN classifiers, and an independent test set $T$ used to evaluate the classifier's generalization performance on unseen data. 

The task of ER active learning is formally defined as follows:

\begin{definition} \label{def:task}
  Provided with the test and validation sets $T$ and $V$, the labeled set $L$ and the unlabeled set (the pool) $U$, active learning iteratively selects a batch of data $Q \subseteq U$ that maximizes a specified criterion given a classifier $h_L$ trained on $L$. At each iteration, once $Q$ is labeled, it is removed from $U$ and added to the labeled set $L$, i.e. $U \gets U \setminus Q$, $L \gets L \cup Q$; finally, a classifier is retrained on the updated set $L$.
\end{definition}

\subsection{Risk Analysis for ER: LearnRisk}
\label{sec:learnrisk}

The risk analysis pipeline operates in three main steps: \textit{Risk feature generation} followed by \textit{Risk model construction} and finally \textit{Risk model training}. 

\subsubsection{Risk feature generation}
This step automatically generates risk features in the form of interpretable rules based on one-sided decision trees. The algorithm ensures that the resulting rule-set is both discriminative, i.e, each rule is highly indicative of one class label over the other; and has a high data coverage, i.e, its validity spans over a subpopulation of the workload.
As opposed to classical settings where a rule is used to label pairs to be equivalent or inequivalent, a risk feature focuses exclusively on one single class.
Consequently, risk features act as indicators of the cases where a classifier's prediction goes against the knowledge embedded in them. An example of such rules is:
\begin{displaymath} \label{eq:rule-example}
  r_i[Year]\neq r_j[Year] \rightarrow inequivalent(r_{i}, r_{j}),
\end{displaymath}	
where $r_i$ denotes a record and $r_i[Year]$ denotes $r_i$'s $Year$ attribute value. 
With this knowledge, a pair predicted as \emph{equivalent} whose two records have different publication years is assumed to have a high risk of being mislabeled. 

\subsubsection{Risk model construction}
Once high-quality features have been generated, the latter are readily available for the risk model to make use of, allowing it to be able to judge a classifier's outputs backing up its decisions with human-friendly explanations. To achieve this goal, LearnRisk, drawing inspiration from investment theory, models each pair's equivalence probability distribution (portfolio reward) as the aggregation of the distributions of its compositional features (stock rewards). 

Practically, the equivalence probability of a pair $d_i$ is modeled by a random variable $p_i$ that follows a normal distribution $\mathcal{N}(\mu_i, \sigma_i^2)$, where $\mu_i$ and $\sigma_i^2$ denote its expectation and variance respectively. Given a set of $m$ risk features ${f_1,f_2,...,f_m}$, let $\mathbf{w}=[w_1,w_2,...,w_m]$ denote their corresponding weight vector. Suppose that $\mathbf{\mu}_F=$ $[\mu_{f_1},$ $\mu_{f_2},$ $\ldots, \mu_{f_m}]^T$ and $\mathbf{\sigma}^2_F=$ $[\sigma_{f_1}^2,$ $\sigma_{f_2}^2,$ $\ldots, \sigma_{f_m}^2]^T$ represent their corresponding expectation and variance vectors respectively, such that $\mathcal{N}(\mu_{f_j},\sigma_{f_j}^2)$ denotes the equivalence probability distribution of the feature $f_j$. Accordingly, $d_i$'s distribution parameters are estimated by:
\begin{displaymath}
  \mu_i = \mathbf{b}_i (\mathbf{w} \circ \mathbf{\mu}_F) \, ; \, \sigma_i^2 = \mathbf{b}_i (\mathbf{w}^2 \circ \mathbf{\sigma}^2_F)
\end{displaymath}
Where $\circ$ represents the element-wise product and $\mathbf{b}_i$ is a one-hot feature vector.

Note that besides one-sided decision rules, LearnRisk also incorporates classifier output as one of the risk features. Provided with the equivalence distribution $p_i$ for $d_i$, its risk is estimated by the metric of Value-at-Risk (VaR)~\cite{tardivo2002value}. Compared with previous simpler alternatives using a single value to represent equivalence probability, LearnRisk can more accurately capture the uncertainty of the label status by a distribution. As observed in investment risk analysis~\cite{artzner1999coherent}, besides expected return, return fluctuation also plays an important role in risk estimation. 

\subsubsection{Risk model training}
Finally, the risk model is trained on a classifier's validation data to optimize a learn-to-rank objective by tuning the risk feature weight parameters ($w_i$) as well as their variances ($\sigma_i^2$). As for their expectations ($\mu_i$), they are considered as prior knowledge, and are estimated from labeled training data. Once trained, the risk model can be used to assess the misclassification risk on an unseen workload labeled by the classifier.

\section{Risk Sampling}
\label{sec:approach}

In AL, each individual iteration can be seen as a standard supervised learning procedure in which a model is fit to labeled data, then the best configuration is selected based on the performance on a disjoint validation set. As shown in Fig.~\ref{fig:framework}, the incorporation of risk analysis as an extra step into the process is therefore fairly straightforward. In this section, we first theoretically derive the optimization model for risk sampling based on the core-set characterization, and then due to its NP-hardness, present a heuristic algorithm for its efficient solution. The notation used throughout this section as well as in Appendix is given in Table~\ref{t:notation}.

\subsection{Optimization Model: Theoretical Derivation} \label{sec:bound}

Based on the core-set characterization for AL presented in~\cite{coreset}, we consider the upper-bound of active learning loss in batch setting defined as
\begin{multline}
     |E_{d,y \sim p_Z}[l(d,y;A_s)]| \\
    \leq \, \Big{|} E_{d,y \sim p_Z}[l(d,y;A_s)] - \frac{1}{n} \sum_{(d_i, y_i) \in D} l(d_i,y_i;A_s) \Big{|} \\ 
    + \, \frac{1}{|s|} \sum_{(d_j,y_j) \in s} l(d_j,y_j;A_s) \\
    + \, \Big{|} \frac{1}{n} \sum_{(d_i, y_i) \in D} l(d_i, y_i; A_s) - \frac{1}{|s|} \sum_{(d_j,y_j) \in s} l(d_j,y_j;A_s) \Big{|}
\end{multline}
\begin{table}[t]
\caption{Notation.}
\label{t:notation} 
\vskip 0.1in
\begin{center}
\begin{small}
\begin{sc}
\begin{tabular}{cl}
    \toprule
    Symbol & Description \\
    \midrule
    $d_i$ & a pair of left and right records \\
          & $\langle \overleftarrow{r_i}, \overrightarrow{r_i} \rangle$ \\
    $y_i$ & pair's label \\ 
    $n_A$ & number of attributes per record \\ 
    $T_{d_i}$ & total number of tokens in pair $d_i$ \\
    $\overleftarrow{a_k}$ ($\overrightarrow{a_k}$) & $k$-th attribute of the left \\
                                                   & (resp. right) record \\
    $w_t^k$ & $t$-th token for attribute $a_k$ \\
    $T_k$ & number of tokens in attribute $a_k$ \\
    $m$ & word embedding dimension \\
    $X_{d_i} \in \mathbb{R}^{T_{d_i} \times m}$ & pair $d_i$'s matrix representation \\
    $X^k \in \mathbb{R}^{T_k \times m}$ & representation of $a_k$ \\ 
    $X^k_t \in \mathbb{R}^m$ & $t$-th token's vector \\  
                             & representation in attribute $a_k$ \\
    \bottomrule
\end{tabular}
\end{sc}
\end{small}
\end{center}
\vskip -0.1in
\end{table}
in which the loss is controlled by the training error of the model on the labeled subset, the generalization error over the full dataset and a term referred to as the core-set loss. Core-set loss is simply the difference between average empirical loss over the set of points which have labels and the average empirical loss over the entire dataset including unlabeled points. Empirically, it is widely observed that DNNs are highly expressive leading to very low training error and they typically generalize well for various classification problems. Hence, the critical part for active learning is the core-set loss. Following this observation, we start off with the core-set loss defined as
\begin{equation} \label{eq:al_erm}
      \Bigg{|}\frac{1}{n} \sum_{(d_i,y_i) \in D} l(d_i,y_i) - \frac{1}{|L \cup Q|} \sum_{(d_j,y_j) \in L \cup Q} l(d_j,y_j)\Bigg{|}
\end{equation}
Where $l$ is the loss for the model trained on $L \cup Q$ ($A_{L \cup Q}$). Informally, given an initial labeled set ($L$) and a budget ($b$), we are trying to find a set of points to query ($Q$), such that the learned model's performance on the labeled subset ($L \cup Q$) and that on the whole dataset ($D$) will be as close as possible. In~\cite{coreset}, it has been shown that provided with a $\lambda${\emph-Lipschitz continuous} convolutional neural network, if a set of balls, denoted by $s$, with radius $\delta_s$ centered at each member of $s$ can cover the entire set $D$, the core-set loss can be bound with the covering radius $\delta_s$ and a term which goes to zero with rate depending solely on $n$. 

The existing core-set characterization applies the global Lipschitz value for all unlabeled points. However, it can be observed that, provided a Lipschitz continuous DNN, the local Lipschitz continuities of unlabeled points are usually not uniform, or their local Lipschitz values may be vastly different.
We implement the AL approach using the classical DeepMatcher model~\cite{deepmatcher} which is built upon recurrent neural networks (RNN). In what follows, we first theoretically establish the Lipschitz continuity of RNN and the DNN model of DeepMatcher, and then derive the optimization model for risk sampling based on non-uniform Lipschitz continuity.

\textbf{Lipschitz Continuity of RNN.} For a generic RNN, we have Lemma~\ref{lem:rnn_lipschitz} on its Lipschitz continuity. We have provided the proofs of the lemmas and theorems in Appendix. 

\begin{lemma} \label{lem:rnn_lipschitz}
    The loss function defined as the 2-norm between one-hot class labels and the Softmax outputs of a stable RNN with $T$ time steps and input dimension $m$, followed by $n_{fc}$ fully connected layers defined over $C$ classes is $\frac{\sqrt{(C {-} 1) \, T \, m}}{C} \, \alpha^{n_{fc} {+} 1}$-Lipschitz.
\end{lemma}

Note that $\alpha$ in Lemma~\ref{lem:rnn_lipschitz} is a bound over the operator norms of all trainable matrices in the RNN and fully connected layers. Although $\alpha$ is in general unbounded, it can be made arbitrarily small without changing the loss function's behavior. Moreover, an RNN is said to be \textit{stable} when the gradients cannot explode, which is only valid when $\alpha < 1$ ~\cite{stableRNN}.
In order to extend the result in Lemma~\ref{lem:rnn_lipschitz} to the DeepMatcher solution for ER, we define a corresponding neural network model, then show that it is Lipschitz continuous in Theorem~\ref{th:dm_lipschitz}.

\begin{definition} \label{def:DNNforER}

\textbf{DNN Model for ER.} The model first embeds each attribute $a_k$ as a sequence of vectors using an embedding matrix $E$ ($X_t^k = E[w_t^k]$). 
Then, each attribute is encoded by a stable RNN into a representation $\mathbf{s_k} \in \mathbb{R}^{m}$ as
\begin{displaymath}
    \mathbf{s_k} = RNN(\mathbf{X^k}).   
\end{displaymath}

Let the attribute similarity layer be defined by a distance function $F_D: \mathbb{R}^{m} \times \mathbb{R}^{m} \xrightarrow{} \mathbb{R}^{m}$. The $k$-th attribute pair similarity $\tilde{s_k}$ between $\overleftarrow{a_k}$ and $\overrightarrow{a_k}$ is then defined as
\begin{displaymath}
    \mathbf{\tilde{s_k}} = F_D(\overleftarrow{\mathbf{s_k}}, \overrightarrow{\mathbf{s_k}}).   
\end{displaymath}

Finally, the classification layer $F_C$ is defined by a fully-connected neural network followed by a Softmax function. The model takes the aggregated pair similarities as input and returns the match probability $p$ by
\begin{displaymath}
    p = F_C(\{\mathbf{\tilde{s_1}}, \dots, \mathbf{\tilde{s}_{n_A}}\}).    
\end{displaymath}
\end{definition}

The model defined in Definition~\ref{def:DNNforER} is consistent with the network structure defined in the RNN variant of DeepMatcher~\cite{deepmatcher}. On its Lipschitz continuity, we have Theorem~\ref{th:dm_lipschitz}.
\begin{theorem} \label{th:dm_lipschitz}
    The loss function defined as the 2-norm between one-hot class labels and the Softmax outputs of an RNN-based ER model as defined in Definition~\ref{def:DNNforER} with input representation dimension $m$ and maximal number of tokens per pair $\hat{T}$ is $\frac{\alpha^{n_{fc} {+} 1}}{2} \sqrt{\hat{T} \, m}$-Lipschitz.
\end{theorem}
\textbf{Optimization Model.}
Based on the Lipschitz continuity of the DNN model for ER, we establish an upper-bound on the core-set loss of active learning in Theorem~\ref{th:upperbound}.

\begin{theorem} \label{th:upperbound}
    Given a dataset $D$ of size $n$ containing a labeled subset $L$ and a Lipschitz continuous classifier, the core-set loss of active learning satisfies the following upper-bound:
    \begin{multline} \label{eq:al_bound}
          \Bigg{|}\frac{1}{n} \sum_{(d_i,y_i) \in D} l(d_i,y_i) - \frac{1}{|L \cup Q|} \sum_{(d_j,y_j) \in L \cup Q} l(d_j,y_j)\Bigg{|} \\ 
          \leq \frac{1}{n} \, \sum_{(d_j,y_j) \in L \cup Q} \sum_{(d_i,y_i) \in C_j} \mathbb{L}_i \, ||X_{d_i} - X_{d_j}||_2
    \end{multline}
    in which $\mathbb{L}_i$ represents its Lipschitz constant for the loss of the model trained on $L \cup Q$, $C_j$ is the $j$-th cluster with $(d_j,y_j) \in L \cup Q$ as its center and $||.||_2$ is the $L_2$ norm.
\end{theorem}

According to Theorem~\ref{th:upperbound}, we define the optimization objective for AL as:
\begin{equation} \label{eq:al_obj}
    \min_{Q}{ \sum_{(d_j, y_j) \in L \cup Q} \sum_{(d_i, y_i) \in C_j} \mathbb{L}_i \, ||X_{d_i} - X_{d_j}||_2}.
\end{equation}
Unfortunately, in (\ref{eq:al_obj}), $\mathbb{L}_i$ is not available prior to the selection of $Q$ and the training of $A_{L \cup Q}$. However, it can be observed that given an unlabeled point, its Lipschitz value depends to a large extent on its misprediction risk. Indeed, if we consider an unlabeled point's misprediction risk as its expected loss, its Lipschitz value can be empirically estimated by
\begin{equation} \label{eq:lipschitz}
  \mathbb{L}'_i = \frac{R_L(d_i)}{\min_{(d_j, y_j) \in L}{||X_{d_i} - X_{d_j}||_2}},
\end{equation}
in which $d_i$ and $d_j$ denote an unlabeled point and a labeled point, respectively. $R_L(d_i)$ denotes the misprediction risk of $d_i$.
This follows straightforwardly from the Lipschitz constant definition for the DNN loss function ($|l(d_i,y_i) \, – \, l(d_j,y_j)| \leq \mathbb{L} \, ||X_{d_i} – X_{d_j}||_2$). 
Since the loss of the labeled pair is assumed to be zero, the loss of the unlabeled pair is estimated via its misprediction risk $R_L(d_i)$.
Therefore, we approximate $\mathbb{L}_i$ with its empirical estimation based on the latest classifier, which is conveniently available as shown in (\ref{eq:lipschitz}). The optimization objective of risk sampling is finally defined as 
\begin{equation} \label{eq:al_obj2}
    \min_{Q}{ \sum_{(d_j, y_j) \in L \cup Q} \sum_{(d_i,y_i) \in C_j} \mathbb{L}'_i \, ||X_{d_i} - X_{d_j}||_2}.
\end{equation}
\subsection{Algorithm} \label{sec:algo}

\begin{algorithm}[tb]
\caption{Weighted fastPAM}
\label{alg:algorithm}
    \KwIn{$D$ \quad \, \ : \, Full data\\
    \qquad \quad $L$ \quad \, \, : \, Initial labeled data\\
    \qquad \quad $b > 0$ \ : \, Query budget
    }
    \KwOut{ Query $Q$.}
    \begin{algorithmic}[1] 
        \STATE Let $Q \gets$ top $b$ points ranked by $\mathbb{L}_i$ 
        \STATE Calculate $TD$ for the initial solution $L \cup Q$\\
        \REPEAT
            \FORALL {$x_j \in D \setminus (L \cup Q)$}
                \STATE $d_j \gets \mathbb{L}_j \cdot d_{nearest}(x_j)$
                \STATE $\Delta TD \gets (0,..,0,-d_j, ..., -d_j)$
                \FORALL {$x_o \neq x_j$}
                    \STATE $d_{oj} \gets d(x_o, x_j)$
                    \IF{$n \in Q$}
                        \STATE Update $\Delta TD_n$
                    \ENDIF
                    \IF{$d_{oj} \leq d_n$}
                        \STATE Update $\Delta TD_i \, for \,  m_i \in Q \setminus \{m_n\}$
                    \ENDIF
                \ENDFOR
                \STATE Save best swap $(\Delta TD^*, m^*, x*)$
            \ENDFOR
            \IF{$\Delta TD^* < 0$}
                \STATE Swap($m^*$,$x^*$)
                \STATE $TD \gets TD + \Delta TD^*$
            \ENDIF
        \UNTIL{$\Delta TD^* \geq 0$}
    \end{algorithmic}
\end{algorithm}

\begin{figure*}[!t]
	\centering
    	\subfloat[][DBLP-Scholar]
	{\includegraphics[width=0.24\linewidth]{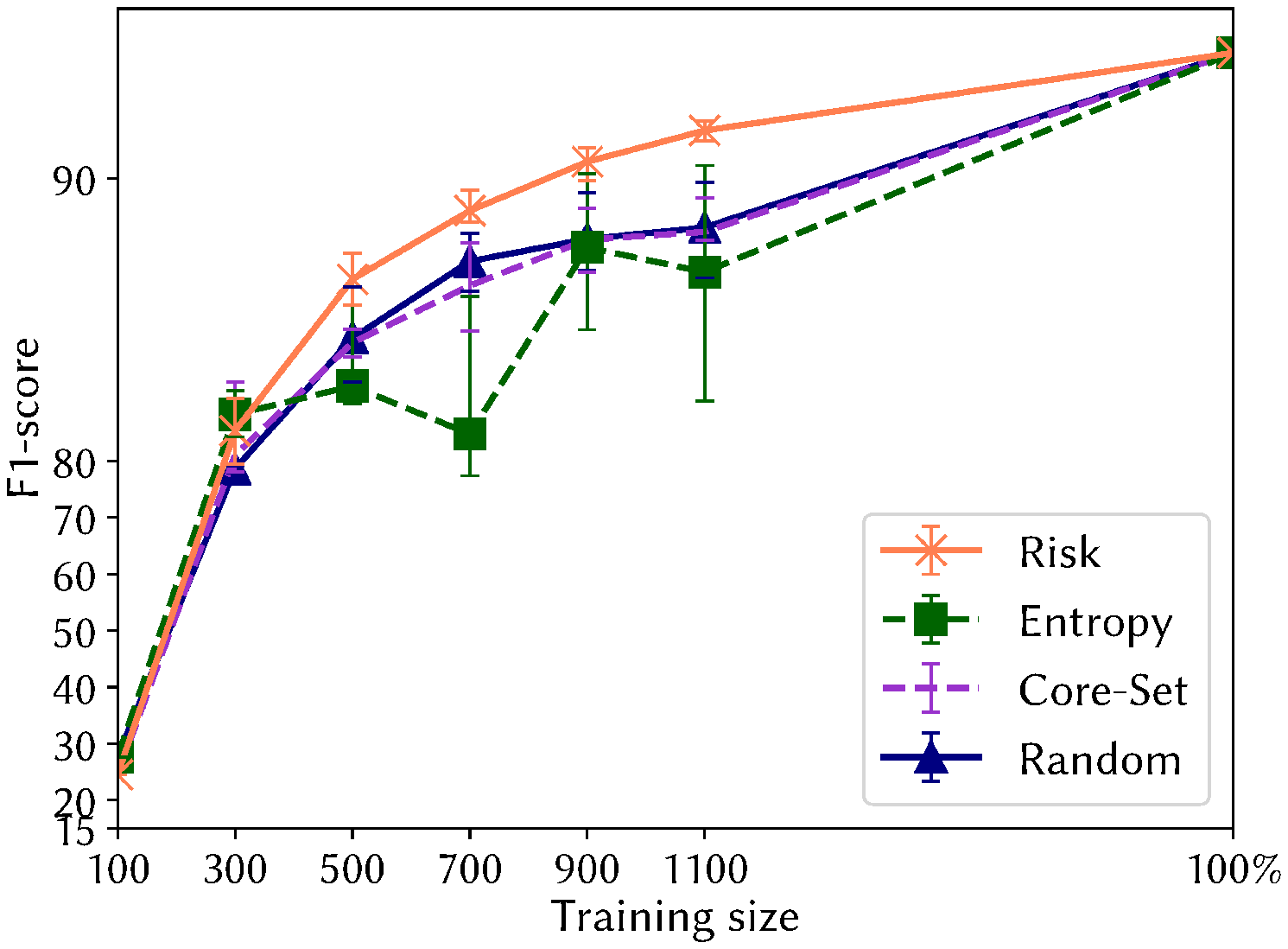} 
	\includegraphics[width=0.24\linewidth]{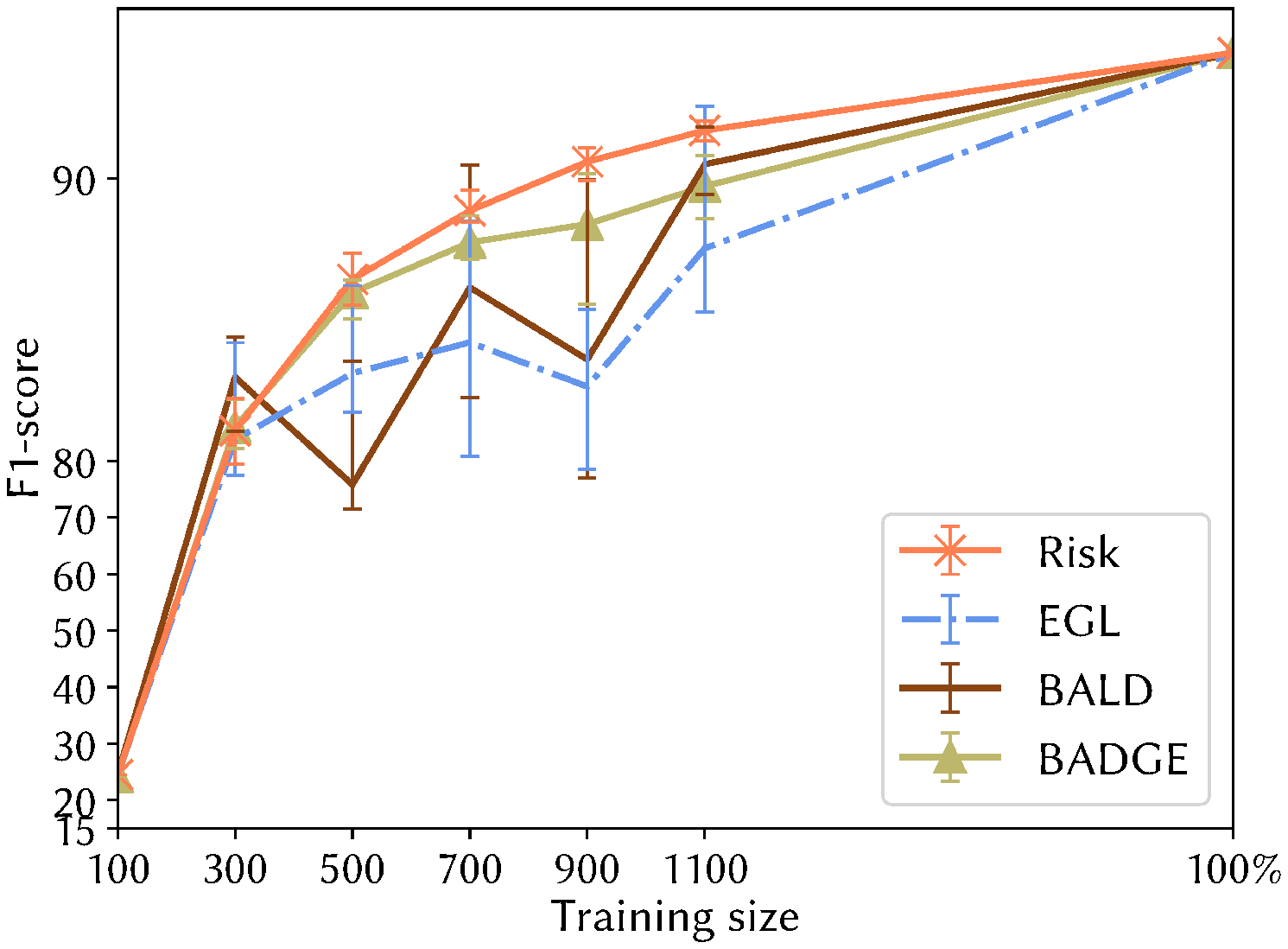}}
    	\subfloat[][Citeseer-DBLP]
	{\includegraphics[width=0.24\linewidth]{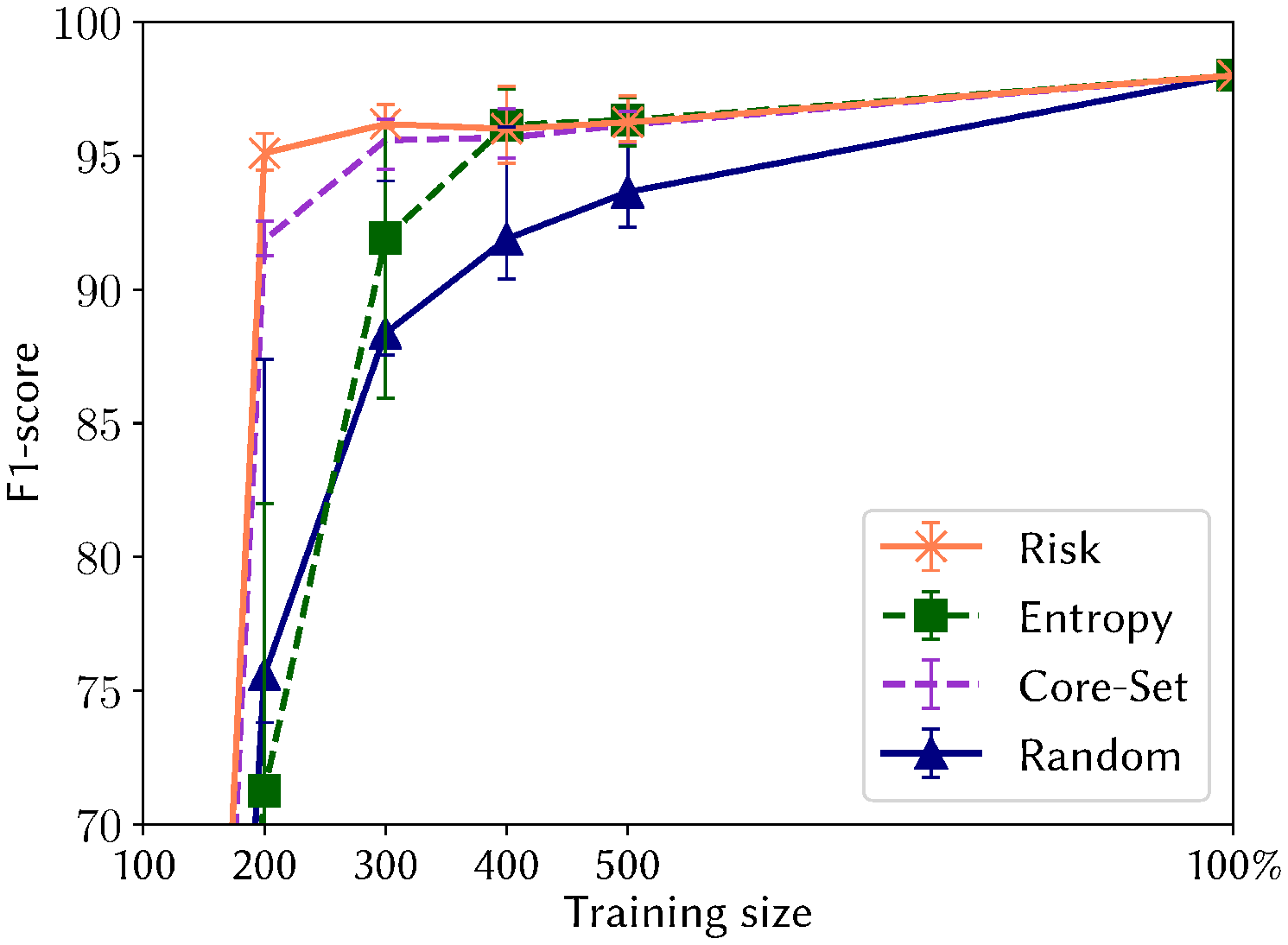} 
	\includegraphics[width=0.24\linewidth]{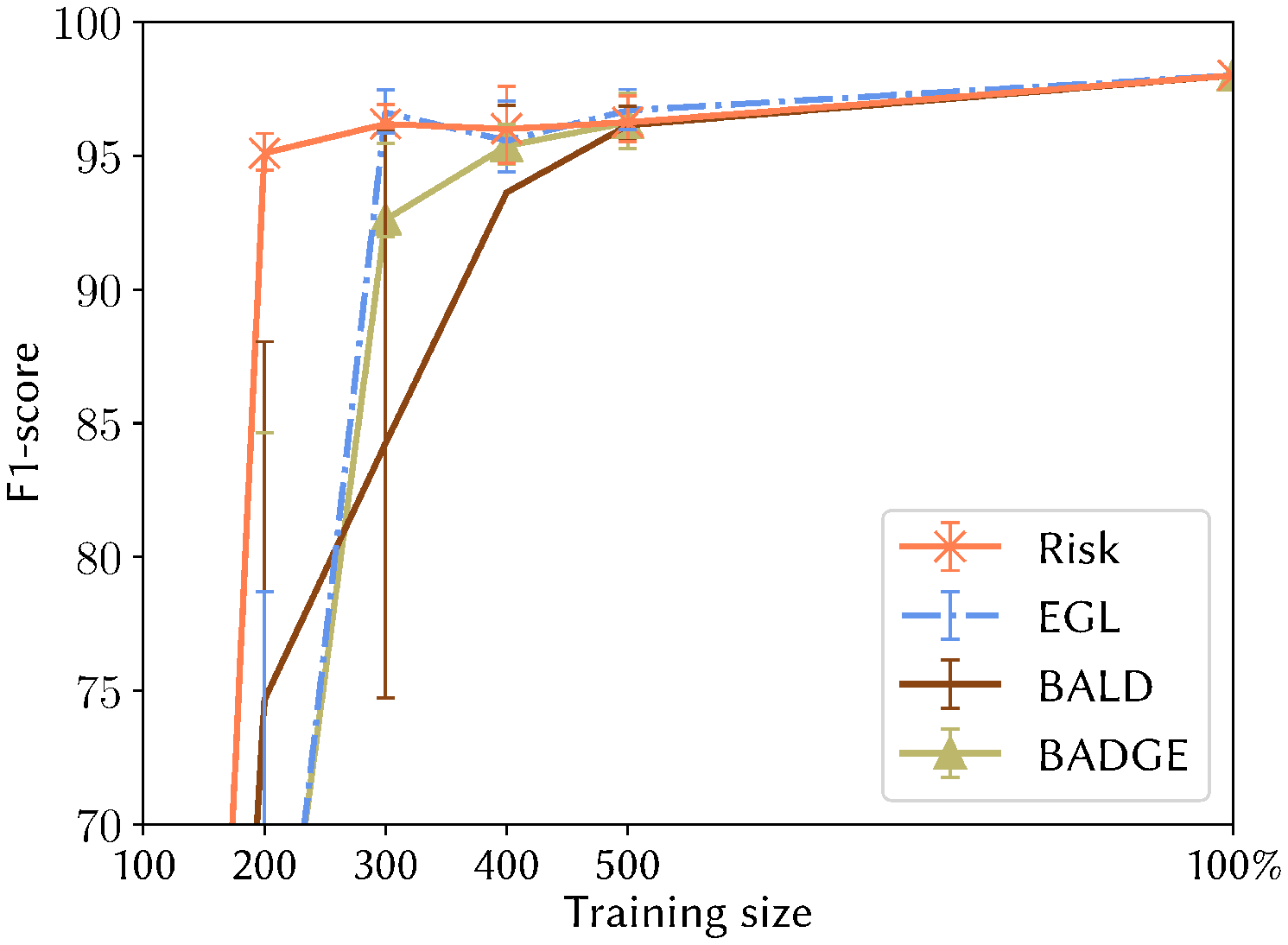}}  \hfill
    	\subfloat[][Abt-Buy]
	{\includegraphics[width=0.24\linewidth]{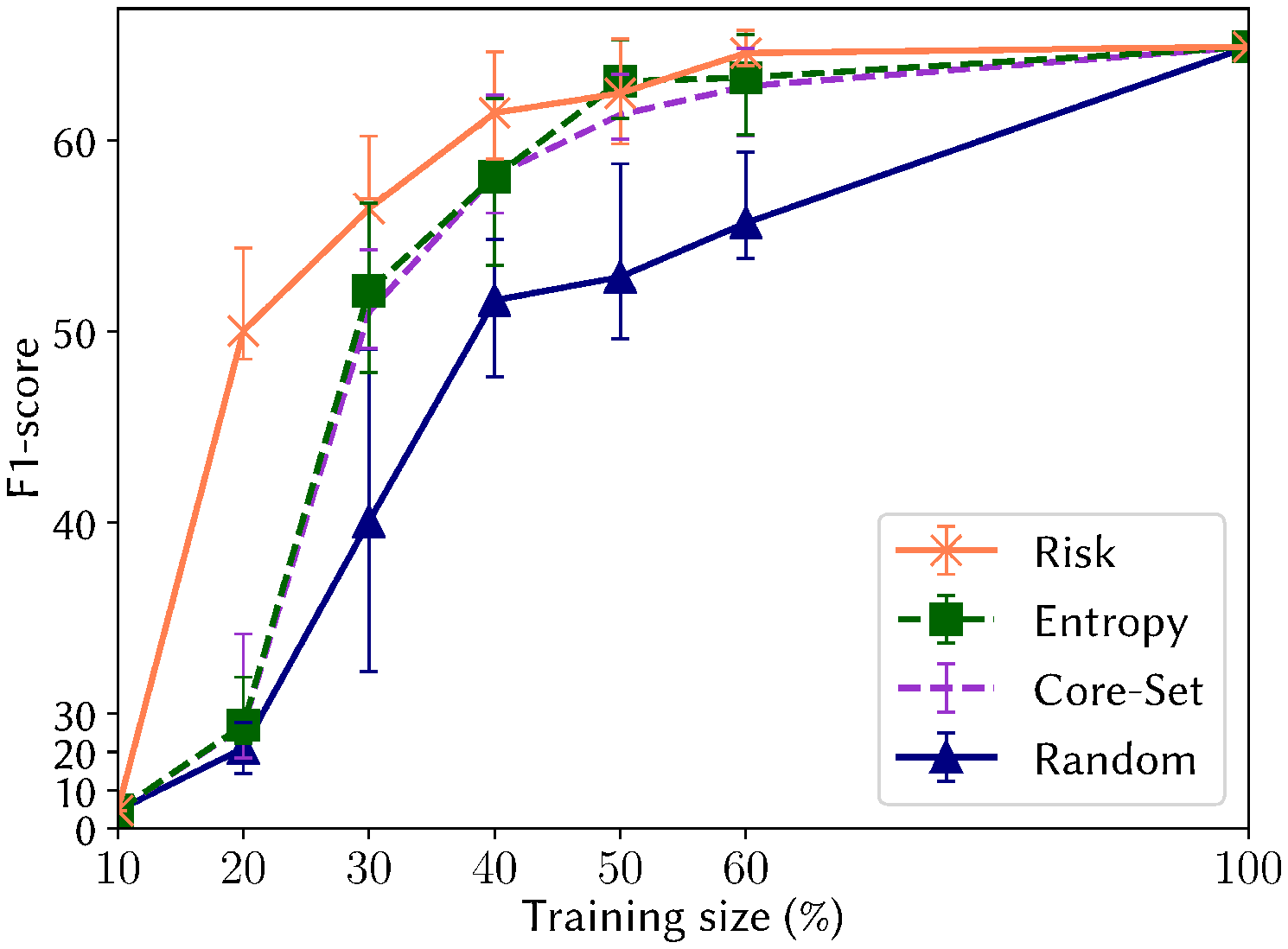} 
	\includegraphics[width=0.24\linewidth]{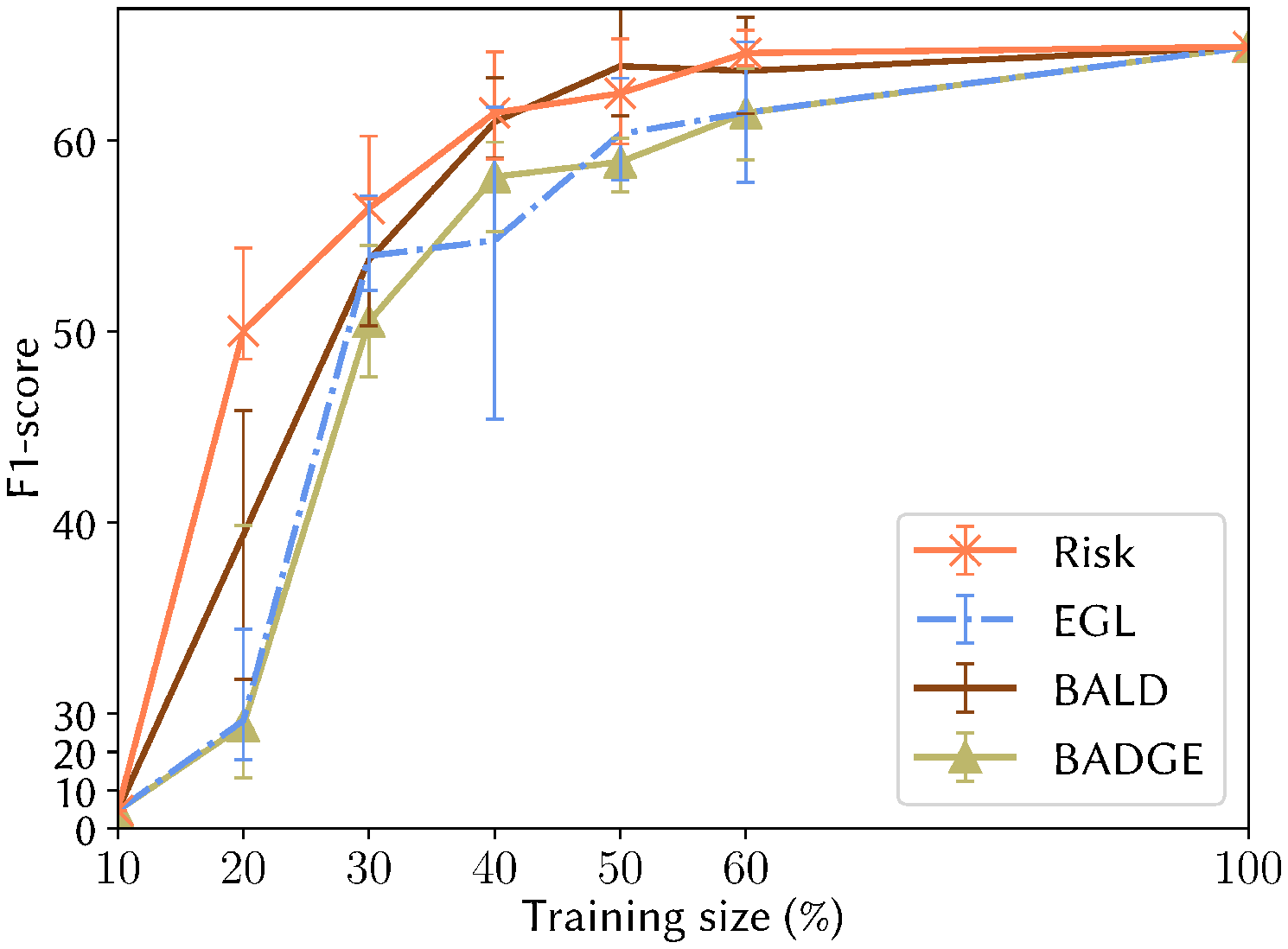}}
        \subfloat[][Songs]
	{\includegraphics[width=0.24\linewidth]{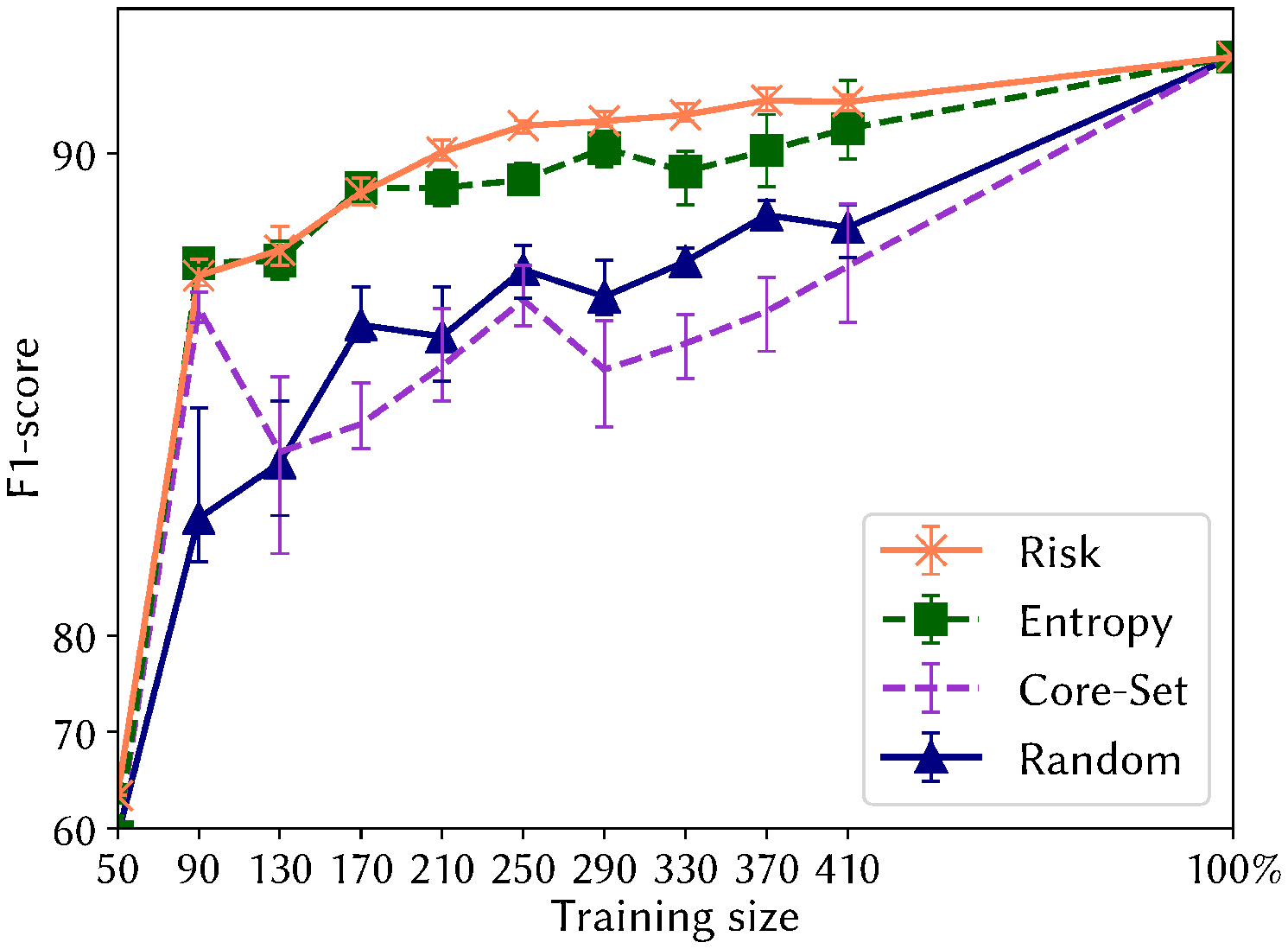} 
	\includegraphics[width=0.24\linewidth]{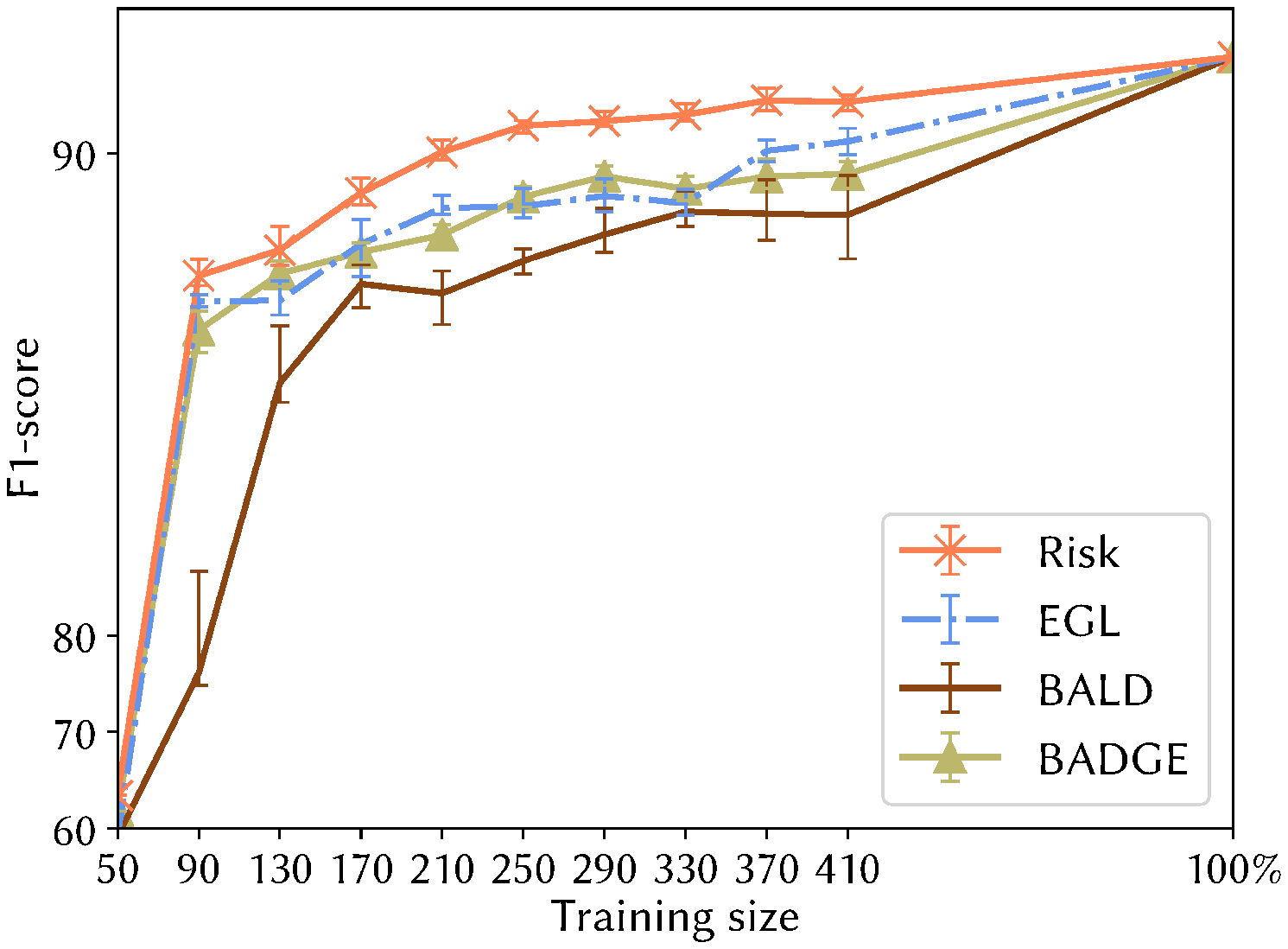}}
	\caption{Comparative Evaluation between AL methods on 4 datasets \textbf{(a-d)}. The comparison on each dataset is split in 2 method groups where \textbf{Risk} denotes the Risk sampling approach. Performance is evaluated by test F1-score per training data size. Error bars indicate the upper and lower quintiles among 10 runs.}
	\label{fig:experiments}
\end{figure*}

Clearly, the optimization problem defined in (\ref{eq:al_obj2}) is a sample-weighted version of the classical k-medoids clustering problem~\cite{pam} with the addition of the weight $\mathbb{L}_i$ for each non-medoid $\mathbf{x_i}$. Given a specified number of clusters $k$, k-medoids aims at finding $k$ clusters where each cluster is centered around a point in the data. Due to its NP-hardness~\cite{nphard}, the classic way to solve the k-medoids problem is via the heuristic Partitioning Around Medoids (PAM) algorithm~\cite{pam}, or its more recent optimized version, namely, fastPAM~\cite{fastpam}. Hence, we adapt the fastPAM algorithm to risk sampling.

In the scenario of risk sampling, the number of clusters is the size of the labeled data in addition to the data to be queried, i.e. $k = |L \cup Q|$. The criterion in (\ref{eq:al_obj2}), represented by the sum of dissimilarities of each point to the medoid of its cluster weighted by its corresponding sample-weight is the \textit{total deviation} objective with Euclidean distance as its dissimilarity measure. For risk sampling, we need to only optimize $Q$ while keeping $L$ fixed. As fastPAM, the proposed algorithm similarly consists of two phases, BUILD and SWAP. To keep $L$ fixed, we force the initial solution to contain $L$ in the BUILD case, and then only allow the points in $Q$ to be swapped out of the solution in the SWAP phase.

The algorithm is sketched in Algorithm~\ref{alg:algorithm}. The first phase generates an initial solution $L \cup Q$ in line 1. After that, the main search loop for phase two is started at line 3. In each iteration, the algorithm will go through candidate points in line 4, calculating the reduction in the total deviation ($\Delta TD$) for each candidate when swapped in place of any non-labeled medoid ($m \notin L$). Lines 7-12 perform the actual calculation w.r.t each medoid and accumulate the values in the $\Delta TD$ vector. The best swap across candidates and medoids is maintained in $(\Delta TD^*, m^*, x*)$ on line 13. The iteration ends by performing the swap between $m^*$ and $x^*$ as long as it provides a decrease in $TD$. Otherwise, the algorithm has converged and $Q$ is returned as the selected query.

The asymptotic complexity of Algorithm~\ref{alg:algorithm} is in the order of $O(b(n-k)^2)$ per iteration in the worst case. With the right cashing of the pairwise distances and the values returned by $n=nearest()$, $d_n=d_{nearest}()$, and $d_s=d_{second}()$; the execution time is monopolized by the nested loops. In our implementation, we opted for a GPU-friendly version of the algorithm by transforming the internal loops into matrix operations and processing the candidates in a batch-wise manner. The execution time can be orders of magnitude faster than the CPU implementation.

\section{Experiments}
\label{sec:experiments}

In this section, we empirically evaluate the performance of risk sampling on real benchmark datasets. It is organized as follows: Subsection~\ref{sec:setting} describes the experimental setting. Subsection~\ref{sec:comparison} presents the comparative evaluation results. Subsection~\ref{sec:sensitivity} evaluates the robustness of risk sampling w.r.t the size of validation data. Finally, Subsection~\ref{sec:efficiency} assesses the runtime efficiency of Algorithm~\ref{alg:algorithm} for risk sampling.

\subsection{Experimental Setting} \label{sec:setting}

Our testbed consists of four datasets from three domains:
\begin{itemize} 
    \item \textbf{Publications.} From this domain we used Citeseer-DBLP\textsuperscript{\ref{usefulstuff}} and DBLP-Scholar\textsuperscript{\ref{deepmatcher}} datasets. We created the Citeseer-DBLP pairs by performing blocking on the raw records to filter out clearly inequivalent pairs and then extracting a random subsample to make up a test dataset of size 10k.

    \item \textbf{Products.} We selected a dataset containing the record pairs from Abt.com and Buy.com online shopping websites\textsuperscript{\ref{deepmatcher}}.

    \item \textbf{Music.} We manually created the Songs dataset from the 1-Million Songs corpus\footnote{\label{usefulstuff}\url{https://sites.google.com/site/anhaidgroup/useful-stuff/data}}, blocked to generate a dataset of size 30k.
\end{itemize} 

We compare risk sampling with the following alternatives:
\begin{enumerate}
    \item \textbf{Random sampling}. The commonly used baseline method which selects points uniformly from the unlabeled set;
    \item \textbf{Maximum Entropy} and \textbf{BALD}~\cite{bald}. Both are based on uncertainty measurement. \textbf{Maximum Entropy} samples points with the highest entropy value, while \textbf{BALD} chooses points that maximize the mutual information with the model parameters;
    \item \textbf{Core-Set}~\cite{coreset}. It is the state-of-the-art \emph{Representativeness}-based approach for DNNs;
    \item \textbf{EGL}~\cite{EGLtext}. The state-of-the-art approach based on \emph{Expected Model Change}, it chooses points that cause the biggest change to the embedding layer parameters;
    \item \textbf{BADGE}~\cite{badge}. A recently proposed approach which trades off between \emph{diversity and uncertainty} by sampling points with diverse gradient embeddings.
\end{enumerate}

These six techniques can provide a good coverage of the existing effective AL approaches for deep models. We built the AL solution upon the hybrid variant of the classical DNN model for ER, DeepMatcher\footnote{\label{deepmatcher}\url{https://github.com/anhaidgroup/deepmatcher/}}.
For the methods that require test-time dropout, we use a dropout rate of 0.2 in the inputs to the RNN module in the embedding contextualization and word aggregation layers. The number of McDropout iterations is set to 100.
Because \textit{EGL} requires two backward passes for each example (each pass assumes a different class label), its application to the full unlabeled set can be very time-consuming. Thus, we randomly sample an unlabeled subset on which EGL-based selection is performed. For \textit{Core-Set}, \textit{BADGE}, and \textit{Risk}, we use the representations of the classifier's penultimate representation layer, prior to the classification layer, for both representations and gradients.

As per Definition~\ref{def:task}, we use a labeled seed set for the initial model training. We provide 100 labeled examples for publications datasets, 50 examples for Songs, and 575 examples (10\% of the unlabeled pool) for Abt-Buy. Similarly, the budget $b$ was chosen to be in a reasonable range w.r.t each specific dataset domain. $b$ cannot be chosen too small that it does not provide enough data for the DNN model, nor can it be too large that more data is labeled than needed. For example, Songs dataset can converge faster with only a few dozens of pairs while Abt-Buy needs a larger budget to show significant improvements. This is true regardless of the AL method applied. We use a budget of 100 examples for publications datasets, 20 examples for Songs and 10\% for Abt-Buy.

To overcome the randomness caused by different model initializations and training data shuffling, we perform 10 training sessions and report the mean test F1-score. For fair comparison, we make sure that all the methods use the same set of model initializations. For the approaches that require access to the classifier (all except \textit{Random}), we use the model with the best validation performance.

\subsection{Comparative Evaluation} \label{sec:comparison}

The evaluation results have been presented in Fig.~\ref{fig:experiments}. Due to the large number of compared methods, we report their performance on each test dataset in two separate sub-figures.

It can be observed that random sampling has the overall lowest performance. 
This confirms the need for active selection. The simple uncertainty method of maximum entropy achieves highly competitive performance on most of the test datasets, e.g. Abt-Buy, Citeseer-DBLP and Songs. While the other uncertainty method of BALD shows slightly higher performance than deterministic maximum entropy on some datasets. However, the improvement is not sufficiently consistent, possibly due to the quality of the MCDropout approximation. It can also be observed that the Core-Set approach can be highly competitive while only considering instance representativeness on most of the test datasets, e.g. Abt-Buy and Citeseer-DBLP. However, purely built upon instance representation, it is not very stable: on Songs, its performance fluctuates wildly. By maximizing the impact on the classifier, EGL is also able to positively impact its performance. On the other hand, making use of gradient information, BADGE was mostly on par with EGL except on DBLP-Scholar, where the gradient-based diversification gave a better and more stable performance.

It is clear that risk sampling is able to consistently increase the classifier's performance across the test datasets. It can be observed that the performance margins between risk sampling and alternative methods are considerable in most cases, especially in earlier iterations (low training sizes). This result clearly demonstrates that exposing the classifier to high-risk examples in an early stage can effectively accelerate training. Coupled with the representativeness achieved by core-set clustering, it is able to maintain an advantage over alternative methods.
Finally, as shown in Fig.~\ref{fig:experiments}, the error bar plots for risk sampling are relatively short, even for the product dataset of Abt-Buy that seems to show high variance overall. This means that the data selected via risk sampling yields less variance in the classifiers across random initializations.
\begin{figure}[t]
	\centering
	    \includegraphics[width=0.48\linewidth]{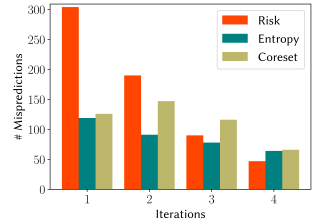}
	\caption{Misprediction selection rate on Abt-Buy.}
	\label{fig:qualitative}
\end{figure}

\textbf{An Illustrative Example.}
The major difference of risk sampling from previous alternatives is the criterion of \textit{misprediction risk}. 
Therefore, we illustrate the efficacy of risk sampling by examining the number of mispredictions in the selected batches on the Abt-Buy dataset, which is the most challenging one. The results are reported in Fig.~\ref{fig:qualitative}.
It can be seen that risk sampling ends up selecting batches dominated by mispredictions. For reference, maximum entropy, which is likely to select mispredictions (since many uncertain points might turn out to be mispredicted), does not pick up as many as risk sampling. The same can be said about the core-set approach which only considers instance representation. The decreasing number of mispredictions throughout iterations is due to the reduction of such cases in the unlabeled pool that we are sampling from. 
Combined with the observation on their comparative performance in the first two iterations, these results clearly indicate that misprediction risk is an informative measure for AL.

\subsection{Robustness w.r.t Validation Data Size} \label{sec:sensitivity}

\begin{figure}[!t]
	\centering
        \subfloat[][DBLP-Scholar]
	{\includegraphics[width=0.48\linewidth]{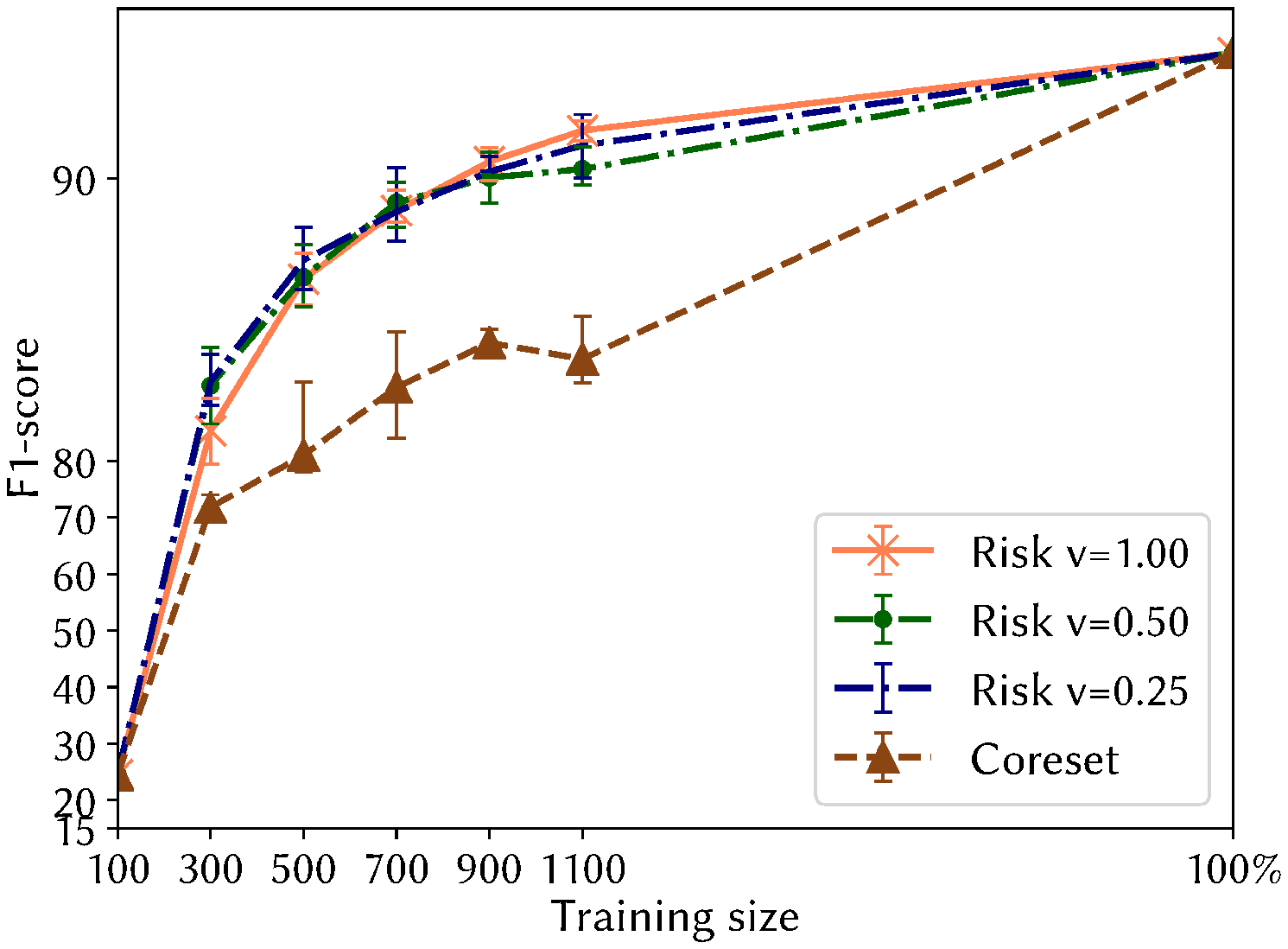}}
        \subfloat[][Citeseer-DBLP]
	{\includegraphics[width=0.48\linewidth]{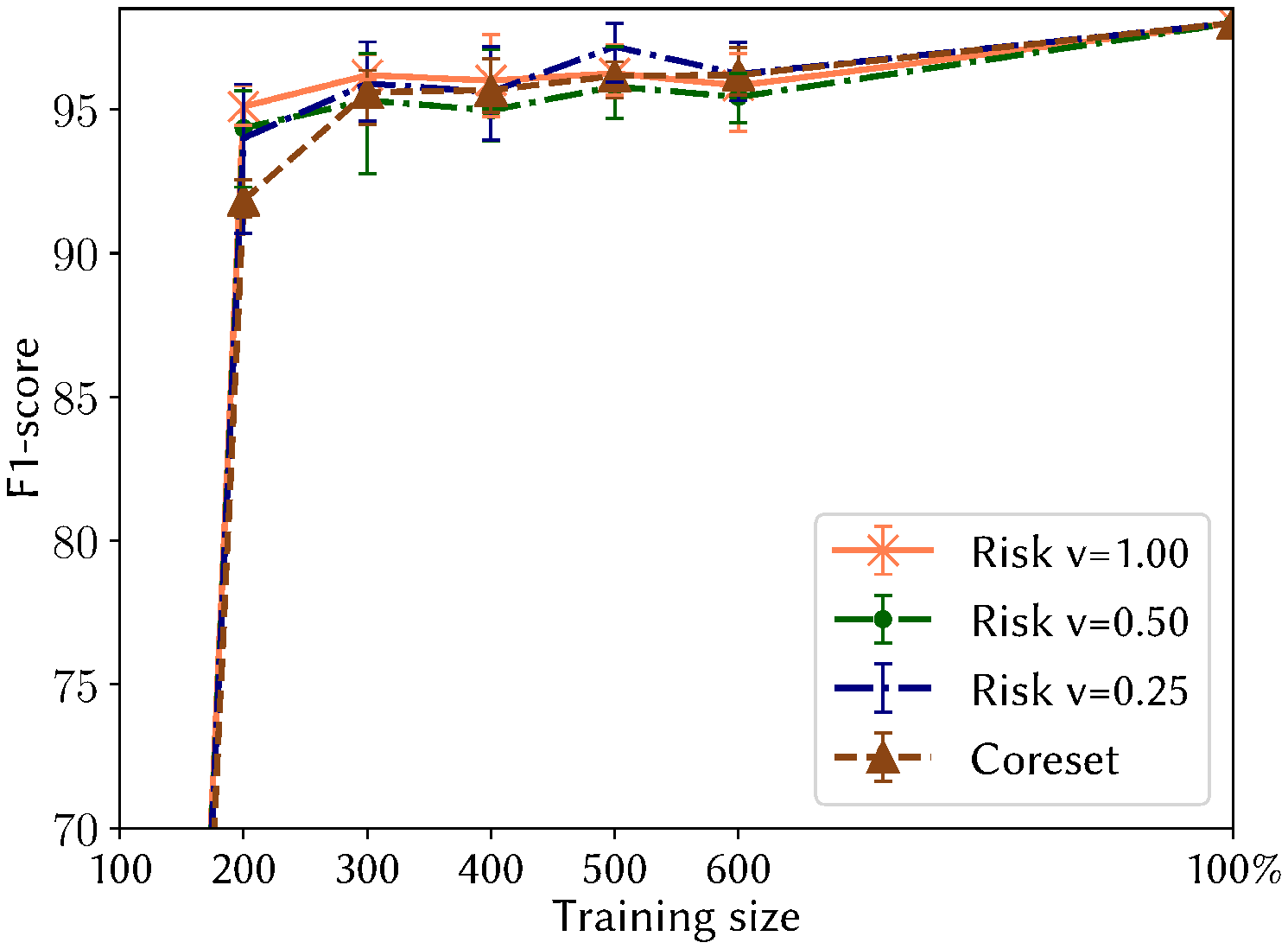}} \hfill
        \subfloat[][Abt-Buy]
	{\includegraphics[width=0.48\linewidth]{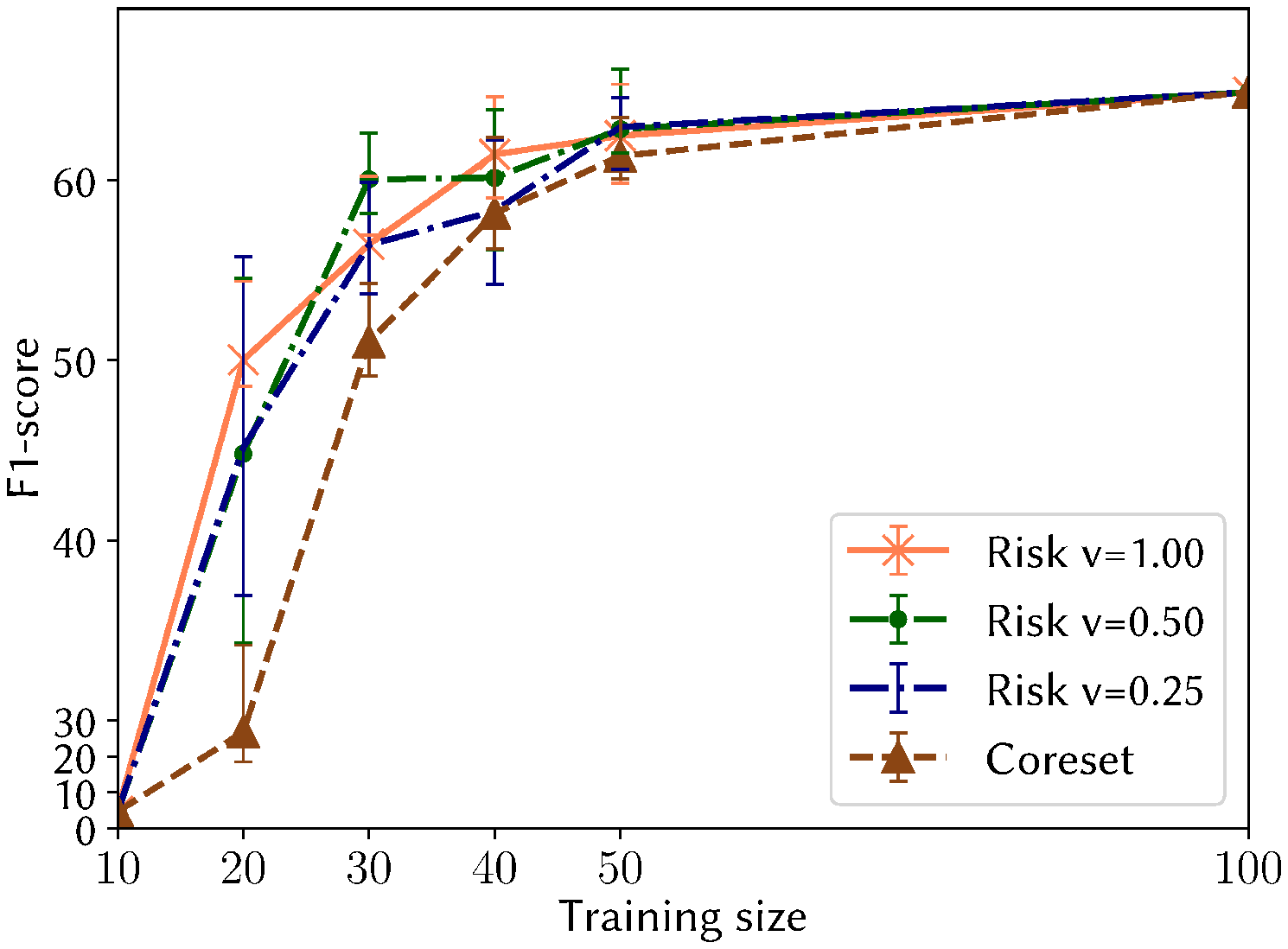}}
        \subfloat[][Songs]
	{\includegraphics[width=0.48\linewidth]{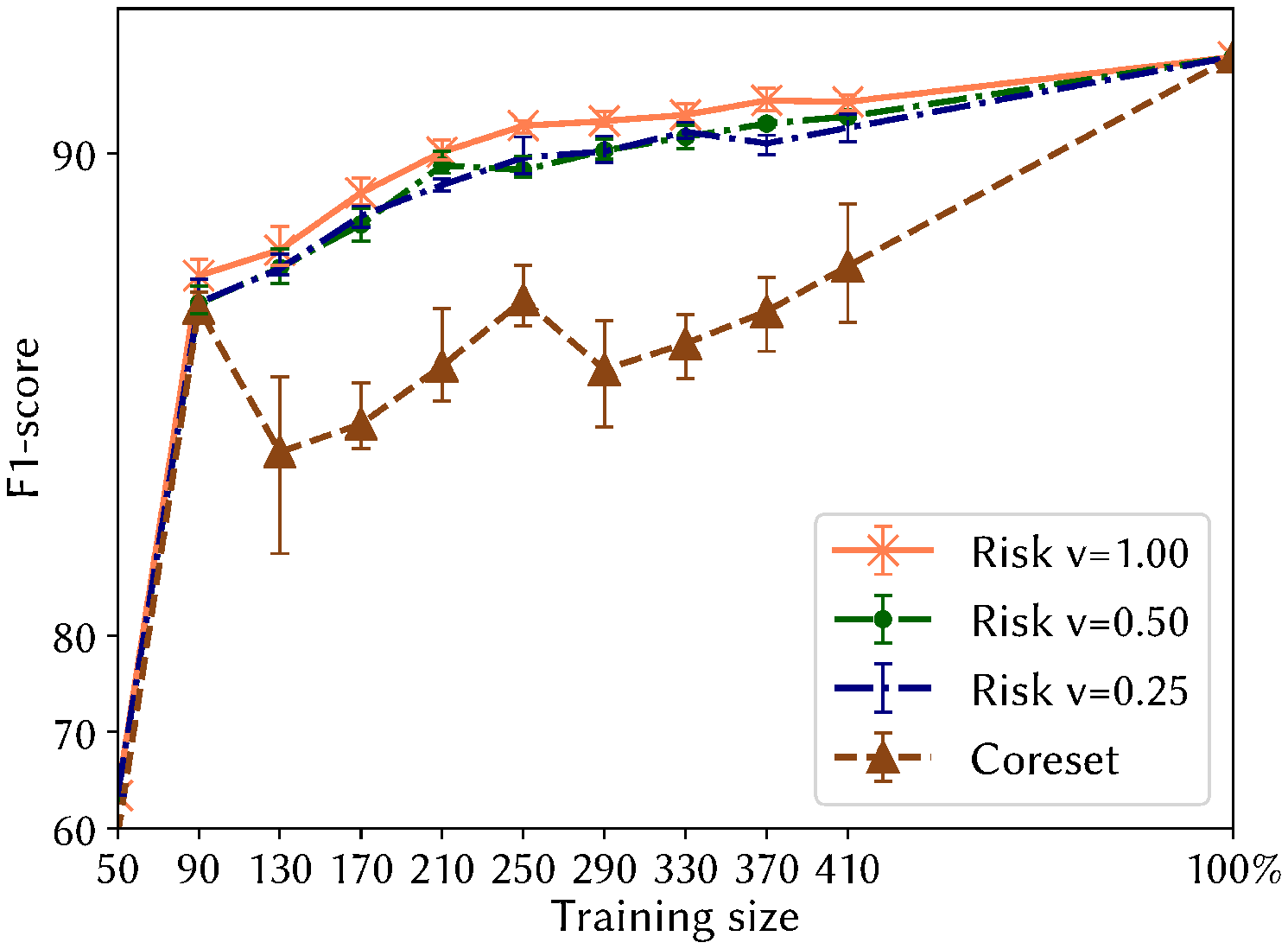}}
	\caption{Robustness Evaluation of Risk Smapling. Evaluates Risk sampling with different validation data ratios (0.5, 0.75, and 1.). Core-set is plotted for performance reference.}
	\label{fig:sensitivity}
\end{figure}

Since risk sampling leverages validation data, we further investigate its performance robustness w.r.t the size of validation data. To this end, we re-run the AL experiment by varying the validation data ratio used for risk training among 0.25, 0.50 and 1. The results on all datasets are presented in Fig.~\ref{fig:sensitivity}. For performance reference, we also plot the result of the core-set approach in the figure. 
It can be observed that the performance of risk sampling is overall very robust across ratios, and it consistently outperforms the core-set approach.  It
is noteworthy that our evaluation results are consistent with those reported in~\cite{RISK}, which showed that the performance of LearnRisk is very robust w.r.t the size of validation data. These experimental results bode well for the application of risk sampling in real scenarios.
\subsection{Risk Sampling Efficiency} \label{sec:efficiency}

\begin{figure}[t]
	\centering
        \subfloat[][Runtime scalability]
	{\label{fig:runtime} \includegraphics[width=0.48\linewidth]{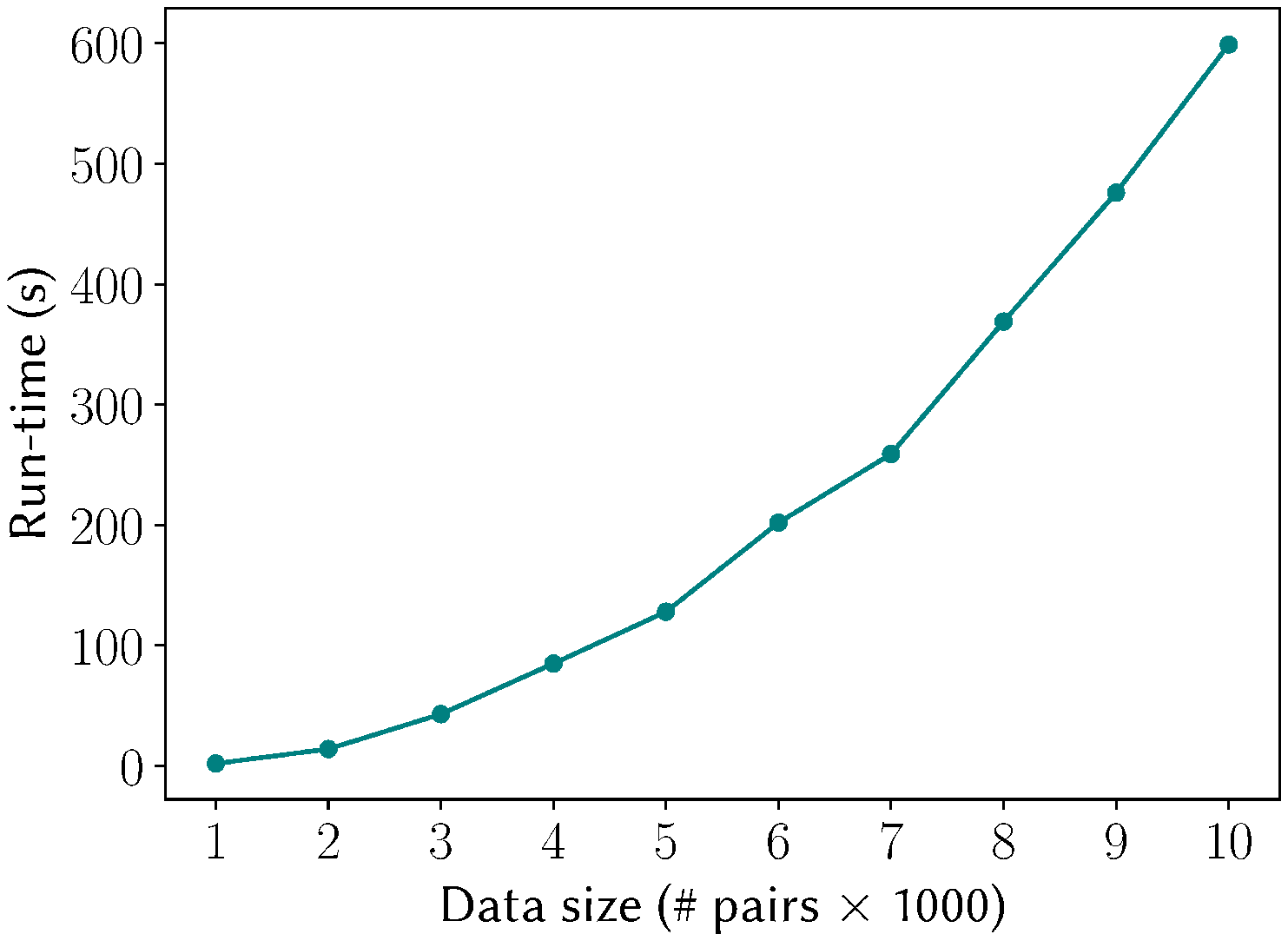}}
        \subfloat[][Convergence scalability]
	{\label{fig:swaps} \includegraphics[width=0.48\linewidth]{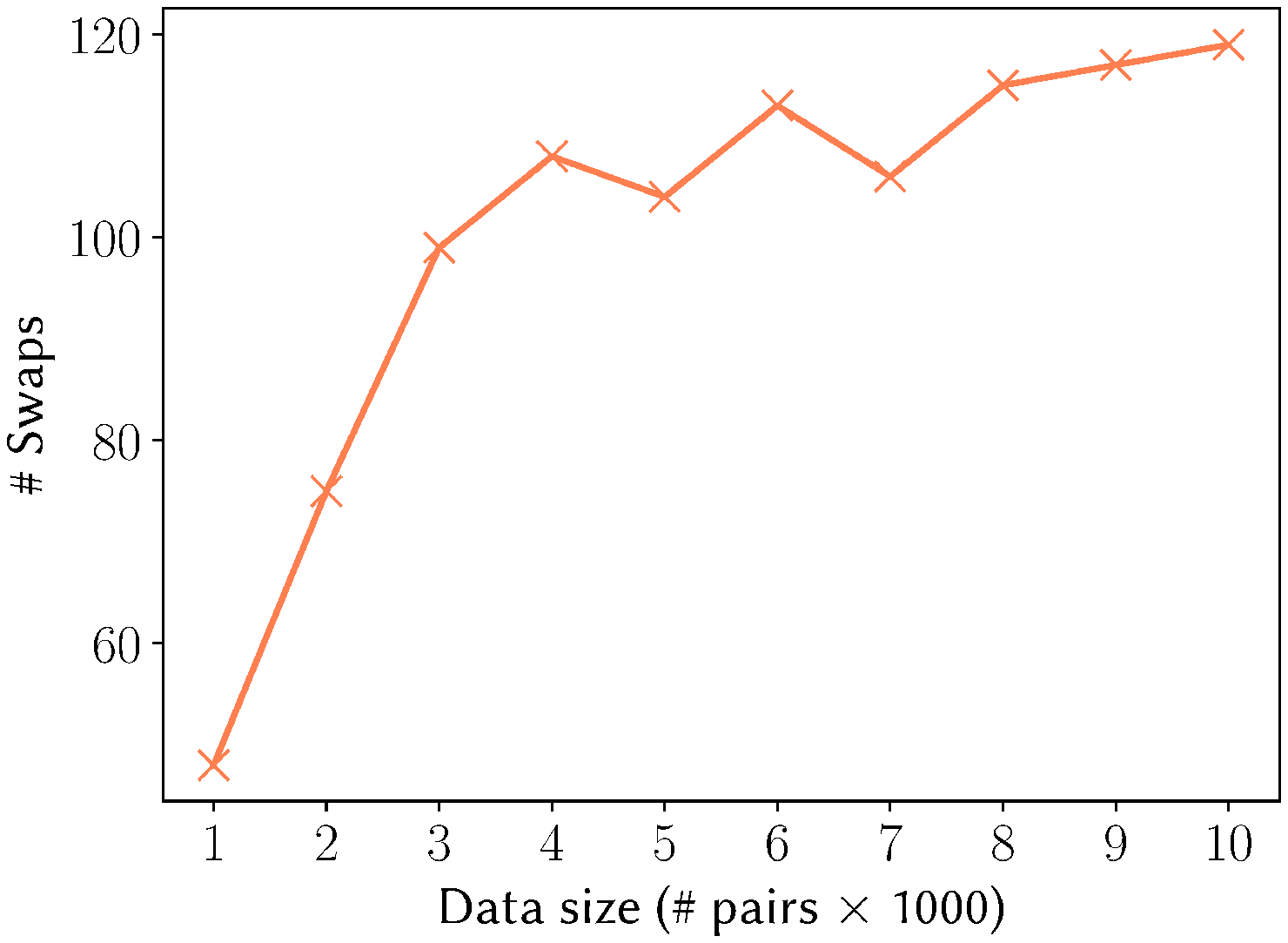}}
	\caption{Risk Sampling scalability.}
    \label{fig:scalability}
\end{figure}

In this section we evaluate the efficiency of the risk sampling algorithm presented in Subsection~\ref{sec:algo}. To this end, we evaluate its scalability w.r.t the total data size ($n$) both in terms of total runtime and number of swaps till convergence. 
We fix the number of clusters $k = 200$ ($|L|=100, |Q|=100$) and variate the data size on the large dataset of DBLP-Scholar using the risk scores and data representations from the first iteration of active learning. The runtimes for the different data sizes are presented in Fig.~\ref{fig:runtime}.
Knowing that the algorithm's time complexity of $O(b(n-k)^2)$ is dependent on $n-k$, it is clear that the combination of a small $k$ ($200$) and a large $n$ ($10000$) still converges in a reasonable time.

Moreover, the plot presenting the number of swaps needed until convergence as a function of data size is given in Fig.~\ref{fig:swaps}. It clearly shows that the number of swaps increases at a slow rate with larger data set size ($n$). Meaning that the execution time is greatly due to the time needed for the search for each swap.

\section{Conclusion}
\label{sec:conclusion}
In this work, we propose a novel strategy of risk sampling for active learning that selects representative points with high misclassification risk for labeling. Built upon the core-set characterization for AL, we theoretically derive an optimization model based on an upper-bound of the core-set loss with non-uniform Lipschitz continuity. Due to the NP-hardness of the defined problem, we then present an efficient algorithm for its solution. Finally, our empirical study has validated the efficacy of the proposed approach. For future work, it is worthy to point out that risk sampling is generally applicable to other classification tasks; their technical solutions however need further investigations.


\bibliography{main}
\bibliographystyle{icml2021}

\newpage
\appendix \label{appendix}

\section{Proof of lemma 1}
We use the following definition of RNN:
\begin{displaymath}
    {\bf{h_t}} = \sigma(W \cdot {\bf{h_{t-1}}} + U \cdot {\bf{x_t}})
\end{displaymath}
s.t. ${\bf{h_0}} = \Phi; U \in \mathbb{R}^{m \times m'}, W \in \mathbb{R}^{m' \times m'}$ and $\sigma$ is an $L_{\sigma}$-Lipschitz activation function.
Note that, the commonly used activation functions for RNNs (ex. tanh) are 1-Lipschitz (i.e. $L_{\sigma} = 1$).
\begin{proof}
    Let $X \in \mathbb{R}^{T \times m}$ be an input sequence of size $T$ (i.e. $X = \{{\bf{x_1}},..., {\bf{x_T}}\}$).
    For two distinct inputs $X$,$X'$ generating hidden states ${\bf{h_t}}$,${\bf{h_t'}} \in \mathbb{R}^{m'}$ respectively, we have:
    \begin{multline*}
        ||\mathbf{h_T} - \mathbf{h_T'}||_p \leq L_{\sigma} \, ||W||_p \, ||\mathbf{h_{T-1}} - \mathbf{h_{T-1}'}||_p \\
        + \, L_{\sigma} \, ||U||_p \, ||\mathbf{x_T} - \mathbf{x_T'}||_p
    \end{multline*}
    By unfolding the right-hand side in the above inequality,
    \begin{multline*}
        ||\mathbf{h_T} - \mathbf{h_T'}||_p \leq L_{\sigma}^T \, ||W||_p^T \, ||\mathbf{h_0} - \mathbf{h_0'}||_p \\
        + \, L_{\sigma}^T \, ||W||_p^{T-1} \, ||U||_p  \, ||\mathbf{x_1} - \mathbf{x_1'}||_p \\
        + \, \dots + L_{\sigma} \, ||U||_p  \, ||\mathbf{x_T} - \mathbf{x_T'}||_p
    \end{multline*}
    For $||U||_p, ||W||_p \leq \alpha$,
    \begin{displaymath}
        ||{\bf{h_T}} - {\bf{h_T'}}||_p \leq \sum_{t=1}^T \alpha^{T-t+1} \, L_{\sigma}^{T-t+1} \, ||{\bf{x_t}} - {\bf{x_t'}}||_p
    \end{displaymath}
    When $p=2$, for an $L_2$-regularized and stable RNN~\cite{stableRNN} ($\alpha \leq 1$) we have $\max_{t \in [1,T]}{\alpha^t} = \alpha$,
    \begin{displaymath}
        ||{\bf{h_T}} - {\bf{h_T'}}||_2 \leq \alpha \, \sum_{t=1}^T ||{\bf{x_t}} - {\bf{x_t'}}||_2
    \end{displaymath}
    Then by applying Cauchy-Schwartz inequality,
    \begin{displaymath}
        ||{\bf{h_T}} - {\bf{h_T'}}||_2 \leq \alpha \sqrt{T} \, ||X - X'||_F
    \end{displaymath}    
    For a fully-connected network module $F_C$ with $n_{fc}$ layers, trainable parameters $w_{fc}$ and $\sigma$ activations, followed by a $C$-class Softmax function~\cite{coreset},
    \begin{multline*}
        ||F_C(\mathbf{h_T}, w_{fc}) - F_C(\mathbf{h_T'}, w_{fc})||_2 \\
        \leq \, \frac{\sqrt{C {-} 1}}{C} \, \alpha^{n_{fc} {+} 1} \, \sqrt{T} \, ||X - X'||_F
    \end{multline*}
    For $w=\{w_{fc}, W, U\}$ and knowing that for a matrix $X \in \mathbb{R}^{n \times m}$: $||X||_F \leq \sqrt{m} \, ||X||_2$,
    \begin{multline*}
        ||RNN(X, w) - RNN(X', w)||_2 \\
        = \, ||F_C({\bf{h_T}}, w_{fc}) - F_C({\bf{h_T'}}, w_{fc})||_2 \\
        \leq \frac{\sqrt{(C {-} 1) \, T \, m}}{C} \, \alpha^{n_{fc} {+} 1} \, ||X {-} X'||_2
    \end{multline*}
    For any fixed $y$, using the reverse triangle inequality we get,
    \begin{multline*}
    |l(X,y,w) - l(X',y,w)| \\
    = \, \big{|} ||RNN(X, w) - y||_2 - ||RNN(X', w) - y||_2 \big{|} \\
    \leq ||RNN(X, w) - RNN(X', w)||_2 \\
    \leq \frac{\sqrt{(C {-} 1) \, T \, m}}{C} \, \alpha^{n_{fc} {+} 1} \, ||X - X'||_2
    \end{multline*}
\end{proof}
\section{Proof of theorem 1}
Here, we study the Lipschitz continuity for the DNN model defined in Definition 2. We suppose a distance function $F_D(\mathbf{\overleftarrow{s_k}}, \mathbf{\overrightarrow{s_k}}) = \big{|} \mathbf{\overleftarrow{s_k}} - \mathbf{\overrightarrow{s_k}} \big{|}$ as used by the DeepMatcher model.
\begin{proof}
    We start with the expression 
    \begin{displaymath}
        ||\mathbf{s_k} - \mathbf{s_k'}||_2 \leq \alpha \sqrt{T_k} \, ||X^k - {X^k}'||_F
    \end{displaymath}
    Let $\tilde{X^k} = \{\overleftarrow{X^k}, \overrightarrow{X^k}\} \in \mathbb{R}^{(\overleftarrow{T_k}+\overrightarrow{T_k}) \times m}$,
    \begin{displaymath}
    \begin{split}
        ||\mathbf{\tilde{s_k}} - \mathbf{\tilde{s_k}'}||_2 & \leq ||\mathbf{\overleftarrow{s_k}} - \mathbf{\overleftarrow{s_k}'}||_2 + ||\mathbf{\overrightarrow{s_k}} - \mathbf{\overrightarrow{s_k}'}||_2 \\
        & \leq \alpha (\sum_{t=1}^{\overleftarrow{T_k}} ||\mathbf{\overleftarrow{x_t}} - \mathbf{\overleftarrow{x_t}'}||_2 + \sum_{t=1}^{\overrightarrow{T_k}} ||\mathbf{\overrightarrow{x_t}} - \mathbf{\overrightarrow{x_t}'}||_2) \\
        & \leq \alpha \sqrt{\overleftarrow{T_k}+\overrightarrow{T_k}} \, ||\tilde{X^k} - \tilde{{X^k}'}||_F
    \end{split}
    \end{displaymath}

    Finally, the classifier module $F_C$ takes in the concatenated similarities $S = \{\mathbf{\tilde{s_k}}\}_{k=1}^{n_A}$. Let $X_d = \{X^k\}_{k=1}^{n_A} \in \mathbb{R}^{T_d \times m}$ be the representation for pair $d$, s.t $T_d = \sum_{k=1}^{n_A} (\overleftarrow{T_k}+\overrightarrow{T_k})$. And let $\hat{T} = \max_{d_i} T_{d_i}$ be the maximal pair length in D. Then, the resulting similarity matrix satisfies,
    \begin{displaymath}
        ||S - S'||_F \leq \alpha \sqrt{\hat{T}} \; ||X_d - X_d'||_F
    \end{displaymath}

    The final expression for the loss function following the same steps as in the proof of Lemma 1 and setting $C = 2$:
    \begin{displaymath}
        |l(d,y,w) - l(d',y,w)| \leq \frac{\alpha^{n_{fc}+1}}{2} \, \sqrt{\hat{T} \, m} \; ||X_d - X_d'||_2
    \end{displaymath}
\end{proof}
\section{Proof of theorem 2}
\begin{proof}
    Let $(d_i,y_i) \in U$, $(d_j, y_j) \in L$ be an unlabeled and a labeled pair respectively. Let $l(d,y)$ be an $\mathbb{L}$-Lipschitz continuous loss function for any pair $d$ with ground-truth label $y$ w.r.t the model $A_{L \cup Q}$ trained on $L \cup Q$. We have:
    \begin{displaymath}
        |l(d_i, y_i) - l(d_j, y_j)| \leq \mathbb{L}_i \, ||X_{d_i} - X_{d_j}||_2
    \end{displaymath}
    
    Where $\mathbb{L}_i$ represents the Lipschitz bound over the slope of the loss landscape between $d_i$ and $d_j$ ($\mathbb{L}_i \leq \mathbb{L}$).
    Let $\{C_1, C_2,\dots, C_{|L \cup Q|}\}$ represent a clustering of $D$ ($D = \bigcup_j C_j$) where each cluster $C_j$ is centered around $(d_j, y_j) \in C_j$. Using triangle inequality and summing over $(d_i, y_j) \in C_j$,
    \begin{multline*}
        \Bigg{|}\sum_{(d_i, y_i) \in C_j} l(d_i, y_i) -  |C_j| \cdot l(d_j, y_j)\Bigg{|} \\
        \leq \, \sum_{(d_i, y_i) \in C_j} \mathbb{L}_i \, ||X_{d_i} - X_{d_j}||_2
    \end{multline*}

    By summing over all clusters $C_j$ and applying triangle inequality, then multiplying both sides by $\frac{1}{n}$,
    \begin{multline*}
        \Bigg{|}\frac{1}{n} \sum_{(d_i, y_i) \in D} l(d_i,y_i) - \frac{1}{n} \sum_{(d_j, y_j) \in L \cup Q} |C_j| \, l(d_j,y_j)\Bigg{|} \\
        \leq \, \frac{1}{n} \, \sum_{(d_j, y_j) \in L \cup Q} \sum_{(d_i, y_i) \in C_j} \mathbb{L}_i \, ||X_{d_i} - X_{d_j}||_2
    \end{multline*}

    Assuming zero loss for labeled data, i.e. $\forall (d_j, y_j) \in L \cup Q: l(d_j, y_j) = 0$, the cluster-weighted loss average and the simple loss average are equal, yielding:
    %
    \begin{multline*}
        \Bigg{|}\frac{1}{n} \sum_{(d_i, y_i) \in D} l(d_i,y_i) - \frac{1}{|L \cup Q|} \sum_{(d_j, y_j) \in L \cup Q} l(d_j,y_j)\Bigg{|} \\
        \leq \, \frac{1}{n} \, \sum_{(d_j, y_j) \in L \cup Q} \sum_{(d_i, y_i) \in C_j} \mathbb{L}_i \, ||X_{d_i} - X_{d_j}||_2
    \end{multline*}

\end{proof}

\end{document}